\pdfoutput=1

\documentclass[11pt]{article}

\usepackage[preprint]{acl}
\usepackage{times}
\usepackage{latexsym}

\usepackage[T1]{fontenc}

\usepackage[utf8]{inputenc}

\usepackage{microtype}

\usepackage{inconsolata}

\usepackage{graphicx}
\usepackage{hyperref}
\usepackage{url}
\usepackage{enumitem}
\usepackage[most]{tcolorbox}
\newcommand{\xhdr}[1]{{\noindent\bfseries #1}.}
\usepackage{booktabs}
\usepackage{booktabs}
\usepackage{colortbl}
\usepackage{caption}
\usepackage{wrapfig}
\usepackage{subfigure}
\usepackage{CJKutf8}
\usepackage{multirow}
\usepackage{amsfonts} 
\definecolor{mypink}{rgb}{.99,.91,.95}
\definecolor{mygreen}{rgb}{.9,.99,.9}
\definecolor{mygray}{gray}{.9}
\definecolor{Blue}{rgb}{0.05,0.05,0.4}
\definecolor{Red}{rgb}{0.4,0.05,0.05}
\newcommand{\bluenote}[1]{\textcolor{Blue}{#1}}
\newcommand{\rednote}[1]{\textcolor{Red}{#1}}

\title{LongWriter-V: Enabling Ultra-Long and High-Fidelity\\ Generation in Vision-Language Models}

\author{ Shangqing Tu$^{1*}$,Yucheng Wang$^{1*}$, Daniel Zhang-Li$^1$, Yushi Bai$^1$,  Jifan Yu$^1$, Yuhao Wu$^2$, \\
\textbf{Lei Hou$^1$, Hui-Qin Liu$^1$, Zhiyuan Liu$^1$, Bin Xu$^1$, Juanzi Li$^1$} \\
$^1$Tsinghua University, $^2$Singapore University of Technology and Design \\
\includegraphics[scale=0.035]{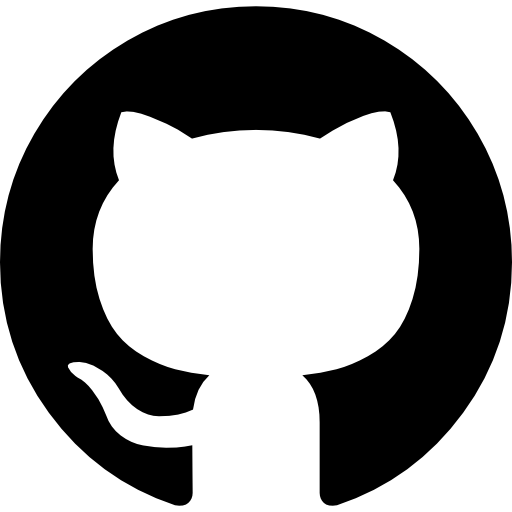} \url{https://github.com/THU-KEG/LongWriter-V} 
\phantom{\thanks{Equal contribution. }}
  }

\begin{document}
\maketitle


\begin{abstract}

Existing Large Vision-Language Models (LVLMs) can process inputs with context lengths up to 128k visual and text tokens, yet they struggle to generate coherent outputs beyond 1,000 words. We find that the primary limitation is the absence of long output examples during supervised fine-tuning (SFT). To tackle this issue, we introduce LongWriter-V-22k, a SFT dataset comprising 22,158 examples, each with multiple input images, an instruction, and corresponding outputs ranging from 0 to 10,000 words. 
Moreover, to achieve long outputs that maintain high-fidelity to the input images, we employ Direct Preference Optimization (DPO) to the SFT model. Given the high cost of collecting human feedback for lengthy outputs (e.g., 3,000 words), we propose IterDPO, which breaks long outputs into segments and uses iterative corrections to form preference pairs with the original outputs.
Additionally, we develop MMLongBench-Write, a benchmark featuring six tasks to evaluate the long-generation capabilities of VLMs. Our 7B parameter model, trained with LongWriter-V-22k and IterDPO, achieves impressive performance on this benchmark, outperforming larger proprietary models like GPT-4o. 

\end{abstract}

\section{Introduction}

Recent advancements in Large Vision-Language Models (LVLMs) have significantly enhanced their capabilities in processing visual and textual inputs~\cite{alayrac2022flamingo,zhang2024vision}.  Notably, there have been substantial breakthroughs in the long-context capabilities of VLMs~\cite{xue2024longvila,shu2024video}. For instance, Qwen2-VL~\cite{wang2024qwen2} can now understand videos up to 20 minutes, with a context window of 32k tokens. This progress has significantly expanded the scope of tasks that VLMs can handle, making them more applicable to real-world scenarios.

\begin{figure*}[t]
    \centering
    \includegraphics[width=\linewidth]{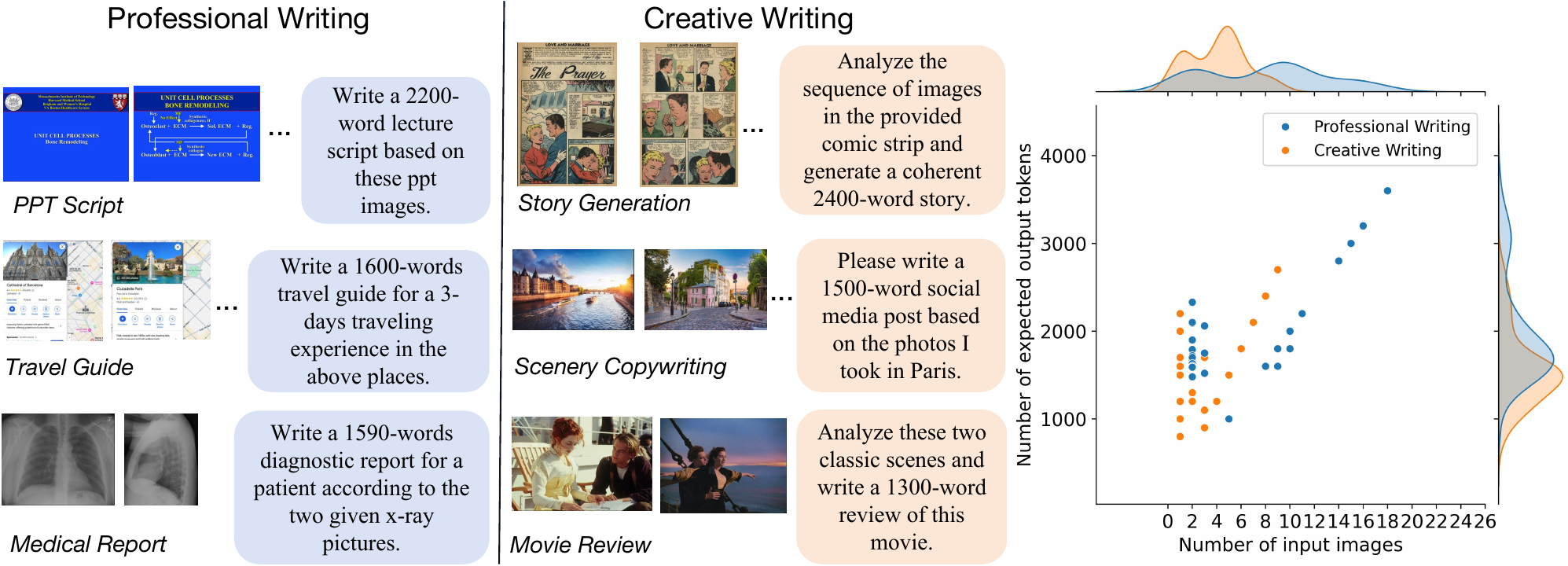}
    \caption{Left: Six examples for each type of task in MMLongBench-Write. They are divided into two categories: professional writing and creative writing. The former requires professional knowledge, while the latter does not. Right: The joint distribution of the number of input images and the expected output length for data in both categories. Most data requires a 1000+ word output with given images, challenging the long-generation capabilities of VLMs.}
    \label{fig:intro}
\end{figure*}

However, despite the increased input context window, the effective output length of VLMs remains limited. To verify this limitation, we collect a benchmark comprising six tasks that require VLMs to generate long texts based on visual inputs (as shown in Figure~\ref{fig:intro}). By adjusting the required output length in the instructions, we found that all existing models struggle to generate outputs exceeding 1,000 words (Section~\ref{sec:preliminary}). In real-world scenarios, such long-output queries are common user demands~\cite{chou2024visionarena}. For example, (1) creative writing tasks may require generating detailed stories or essays based on visual prompts~\cite{hong2023visual}, and (2) professional writing tasks may involve writing comprehensive reports or analyses from visual data~\cite{hartsock2024vision}. To meet these practical needs, it is essential to enhance the long-output capabilities of VLMs.

To investigate the reasons behind the limited long-output capability of VLMs, we are inspired by the LongWriter~\cite{bai2024longwriter}, which adjusts the output length distribution of the supervised fine-tuning (SFT) data to observe changes in model output length. Our experiments revealed that the proportion of long-output examples in the SFT data determines the model's output length. This finding explains why VLMs typically have an output length limit of around 1,000 words. Existing visual instruction tuning datasets~\cite{schuhmann2022laion}, such as LLaVA~\cite{liu2024visual}, mainly contain tasks like grounding~\cite{liu2024grounding} and caption generation~\cite{wang2022git}, with most outputs being less than 300 words~\cite{lin2014microsoft}.

To fill the gap, we select long-output instruction-image pairs from MMEvol~\cite{luo2024mmevol} as inputs. In addition to single-image inputs, we also constructed other forms of data, including multi-image inputs and backtranslated instructions~\cite{wang2024weaver}, to enrich the diversity of the input data. To generate long outputs, we propose a plan-and-write approach: LongWriter-Agent-V. This method involves providing input images and writing instructions to GPT-4o to first generate an outline and then sequentially write the text in segments. Through this approach, we collect LongWrite-V-22k, a dataset of 22k long-output examples.

Using LongWrite-V-22k for SFT, the output length of Qwen2.5-VL-7B-Instruct~\cite{Qwen2.5-VL} can be extended beyond 3,000 words. However, longer outputs often introduce issues such as repetition and hallucination~\cite{favero2024multi}. To improve the fidelity of long outputs, we adapted the approach from RLHF-V~\cite{yu2024rlhf}, where human experts revise the model's outputs to form preference pairs for Direct Preference Optimization (DPO). Since traditional DPO~\cite{rafailov2024direct} is typically performed on short texts of around 300 words, and LongWriter-V's output length can exceed 3,000 words, the annotation cost is extremely high. To enhance the efficiency of preference data utilization, we proposed IterDPO, which divides long outputs into N segments, treating each segment's revision as a preference pair. This method allows the model to learn fine-grained human corrections for each segment and effectively multiplies the use of a single long-output preference pair by N times. Through LongWriter-V-22k SFT and IterDPO, our 7B model achieves impressive performance in both output length and quality, surpassing powerful VLMs like GPT-4o.

In summary, our contributions are as follows:
\begin{itemize}
\item We construct MMLongBench-Write to evaluate the long-output capabilities of VLMs and find that the output length limit of existing VLMs is around 1,000 words.
\item We collect the SFT dataset LongWrite-V-22k, enabling VLMs for 3,000+ word generation.
\item We propose IterDPO, which effectively improves the text quality of long-output VLM.
\end{itemize}

\section{Preliminaries}
\label{sec:preliminary}
In the preliminary experiments, we first collect MMLongBench-Write, a benchmark with visual inputs and long-output requirements. Then, we conduct an evaluation on this benchmark to explore the maximum output length of VLMs.  Besides, we reveal that the main reason for bounded output length lies in the length distribution of SFT data.

\xhdr{MMLongBench-Write}
The ability to write long texts based on visual inputs is a fundamental skill in various real-world applications and can be broadly categorized into professional writing and creative writing, depending on whether specialized knowledge is required~\cite{taavitsainen2000conventions}. 
To evaluate how well that VLMs master the two skills, we design three specific tasks for each skill. For each task, we curate 20 representative instructions with input images as test data. To ensure diversity, half instructions are in English and half are in Chinese. Figure~\ref{fig:intro} shows six examples of the benchmark and data distribution. It highlights that professional writing tasks typically involve more input images and require longer output lengths. 


\xhdr{LongWrite-V-Ruler}
To explore the maximum output length of VLMs, we select 8 examples from MMLongBench-Write benchmark, with four samples in English and four in Chinese. As depicted in Figure~\ref{fig:intro}, each instruction is in the form of \textit{"Write an $L$-word article for the given pictures"}. We construct a diverse test set by changing the length requirement $L$. This test set uses $L \in \{500, 1000, 2000, 4000\}$, which consists of 32 test prompts in total.

\begin{figure}[t]
    \centering
    \includegraphics[width=\linewidth]{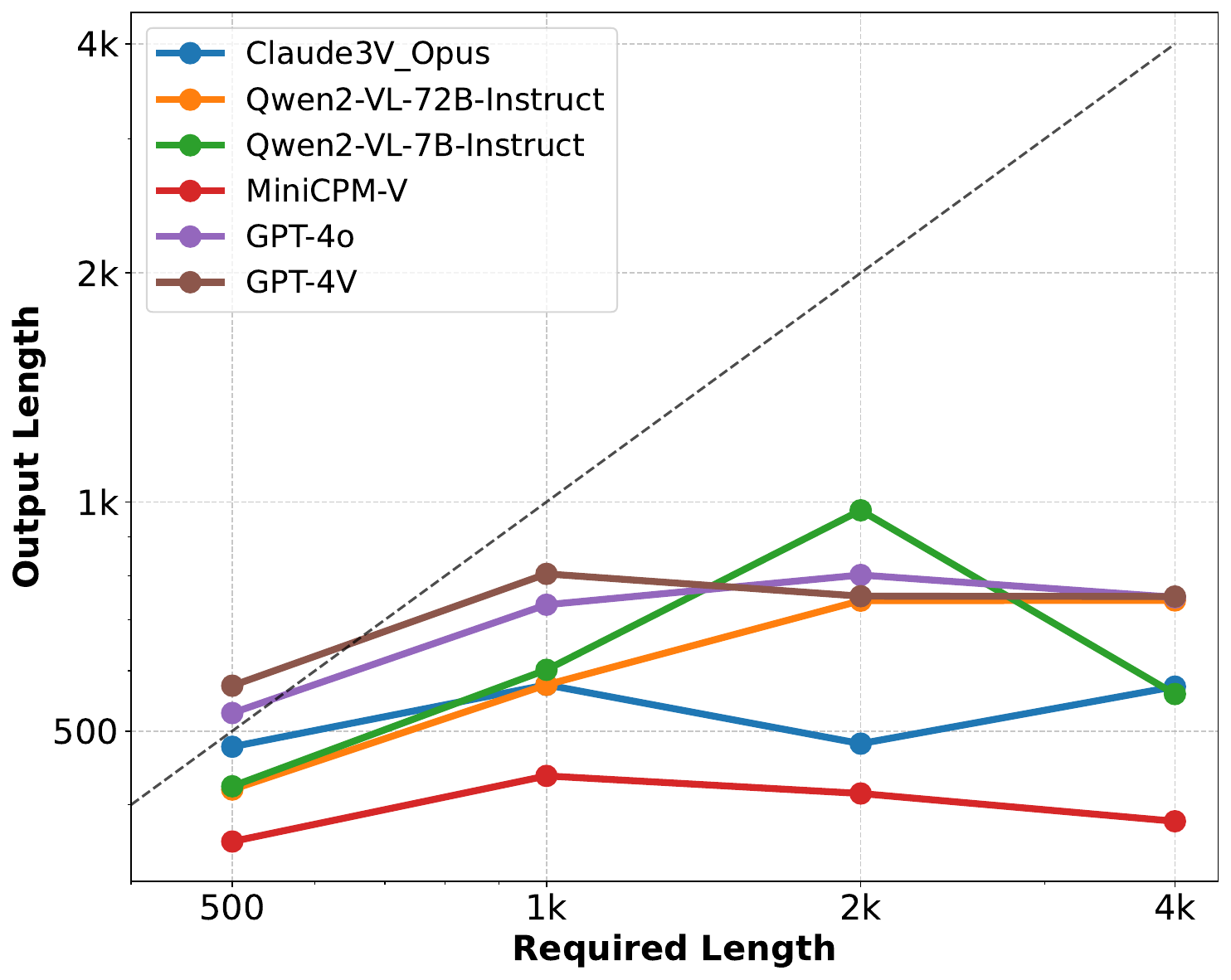}
    \caption{LongWriter-V-Ruler test across different output length requirements. The horizontal line show the overall upper bound for current VLMs.}
    \label{fig:ruler_res}
\end{figure}

\xhdr{Evaluation Result}
We conduct the LongWrite-V-Ruler test on three open-source VLMs and three proprietary models. In Figure~\ref{fig:ruler_res}, we plot the required output length (x-axis) and the corresponding average output length (y-axis) for 12 instructions. We can observe that there exists an upper bound of 1000 output length for all models.

\begin{figure}[t]
    \centering
    \includegraphics[width=\linewidth]{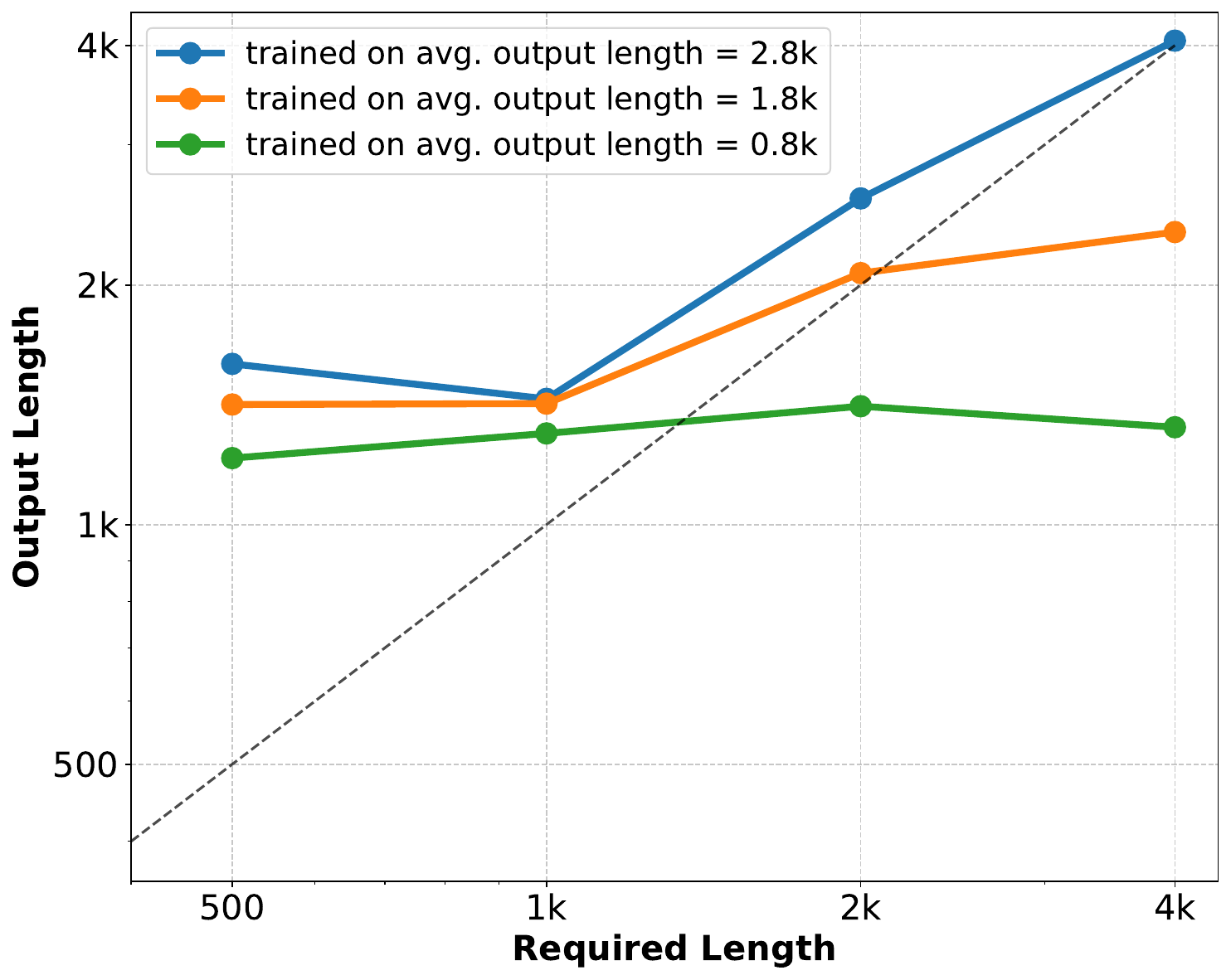}
    \caption{LongWriter-V-Ruler test for Qwen2-VL-7B-Instruct trained on 10k SFT data samples with different average output lengths. }
    \label{fig:ruler_res_sft}
\end{figure}

\xhdr{Preliminary Experiment}
As the controlled experiments in LongWriter~\cite{bai2024longwriter} has revealed that the maximum output length of LLM is correlated with the maximum output length of SFT data, we further explore how the average output length of SFT data can influence the long-generation capabilities of VLM. We fine-tune Qwen2-VL-7B-Instruct~\cite{wang2024qwen2} on three visual instruction datasets sampled from our final SFT data. Each dataset has 10k examples with different average output length respectively (0.8k, 1.8k and 2.8k). Figure~\ref{fig:ruler_res_sft} shows the trained models' performance on LongWrite-V-Ruler, we observe that the model's maximum output length increases with the average output length of SFT data. Besides, we find that \textbf{the number of long-output examples} is crucial for extending the output length of VLMs. For example, the training set with an average length of 1.8k contains 1\% data exceeding 4k output length, but the model trained on it fails to generate 4k tokens (orange line). In contrast, the model trained with 21\% data exceeding 4k output length is able to do that (blue line). This result indicates that the main reason that limits the VLM's output length is lack of enough long-output examples in SFT data.


\section{LongWriter-V: Data and Training}
In this section, we will introduce the data collection and training process for unlocking the long generation capability of vision-language models.

\begin{figure*}[t]
    \centering
    \includegraphics[width=\linewidth]{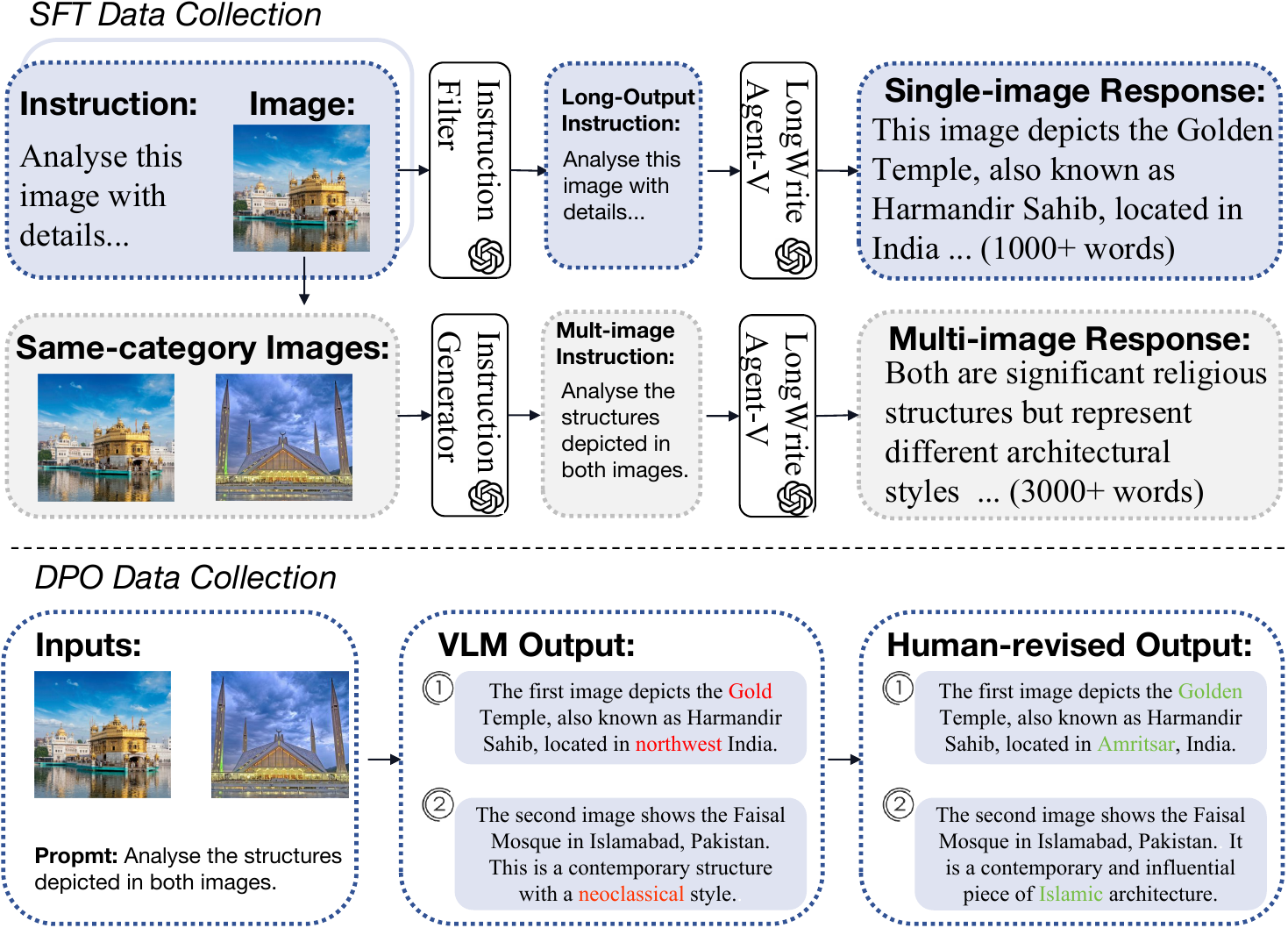}
    \caption{SFT and DPO data collection pipeline of LongWriter-V. The SFT data includes both single-image and multi-image input for long text output. The DPO data contains human revision over each paragraph of VLM's long output. We conduct iterative direct preference optimization to learn the fine-grained human feedback.}
    \label{fig:pipeline}
\end{figure*}

\subsection{Data Collection}
\label{sec:data_collection}
Figure~\ref{fig:pipeline} depicts the overall pipeline of our data collection process, which consists of two phases: SFT and DPO data collection. 

\subsubsection{SFT Data Collection}
\label{sec:sft_data_collection}
Existing VLMs fail to directly generate texts exceeding 1k tokens, so we develop a two-stage method to generate long texts as SFT data.

\xhdr{LongWrite Agent-V} Before introducing our method, we first formalize the task's objective. Given several input images $v$ and an user instruction $x$, our goal is to generate a text $y^*$ that aligned with user's length and quality requirements:
\begin{equation}
   y^* = \mathop{\arg\max}\limits_{y}(s_l(y) + s_q(y) )P_{\theta}(y|v,x) 
\end{equation}
where $s_l$ and $s_q$ is the scoring function for judging the length and quality of the output, respectively. $P_{\theta}$ is the function representing the end-to-end solution, while existing VLMs may not be directly applied as their maximum output lengths are below 1k.
To utilize off-the-shelf VLMs, we propose a two-stage method for generating long texts. Inspired by the plan-and-write method from LongWriter~\cite{bai2024longwriter}, we first prompt the VLM to generate an outline $o$ that structures the output, plans the content, and specifies the word count for each paragraph. This outline breaks down the complex long-output task into manageable sub-tasks. Next, we use the VLM to fill in each paragraph and concatenate them to form the final output:
\begin{equation}
y^* = \mathop{\arg\max}\limits_{o}P_{1}(o|v,x)  \mathop{\arg\max}\limits_{y}(s_l(y) + s_q(y) )P_{2}
\end{equation}
\begin{equation}
P_{2}(y|v,x,o) = \prod \limits_{i=0}^n p(y_i|v,x,o_i,y_{<i})
\end{equation}
where $P_{1}$ is the modeling function for first stage, which takes input images and instruction to write an n-paragraph outline $o = \{o_{i}, i =1,...,n\}$. $P_{2}$ refers to the second stage, where the VLM outputs the content $y_i$ paragraph by paragraph based on the input information, outline $o_{i}$ and previous paragraphs $y_{<i}$. In practice, we design two detailed prompts for guiding VLM to implement the two stages, which are listed in Appendix~\ref{app:prompt_lav}.


\xhdr{Visual Instruction Collection}
To collect long-output visual instructions for SFT, we choose MMEvol~\cite{luo2024mmevol} as our primary data source. MMEvol is a large-scale, open-domain dataset containing 480k image-text instruction pairs, sourced from diverse datasets such as LLaVA-Instruct~\cite{liu2024visual} and ShareGPT4V~\cite{chen2024sharegpt4v}. However, the average output length in this dataset is relatively short (54.85 tokens), necessitating a filtering process to identify long-output instructions. We first check the original response length of e ach example and select those with output length over 128, yielding 55,835 valid data. Next, we utilized GPT-4o to verify whether each instruction genuinely requires a long output and whether the associated image was sufficiently relevant to the instruction. Finally, we get 8,115 single-image instructions.


\xhdr{Multi-image Instruction Generation} As the original data in MMEvol only has one image for each instruction, we synthesize some multi-image instructions to increase the diversity of SFT data.  We select three subsets of MMEvol: wikiart, web-landmark, web-celebrity. Each subset contains hundreds of images in the same category. For example, images in web-landmark are all landmark pictures taken from different world attractions. We randomly sample 2 or 4 same-category images and then ask gpt-4o to generate an instruction that require long output for these images. We obtain 6,313 multi-image instructions in this way. Apart from synthetic data, we also collect natural multi-image data from an open-source PPT dataset, Zenodo10K ~\cite{zheng2025pptagent}. We transform these slides into images to use them as visual inputs and set the instruction as "Write a lecture script for these slides". We choose those slides that has at least 2 pages and at most 30 pages, resulting in 7,730 data.

\xhdr{Backtranslation}
Through above processes, we collect 22,158 single-image and multi-image instructions in total. Using the LongWrite Agent-V pipeline, we generate long output for each visual instruction as SFT data. We call this training data LongWrite-V-22k.  But most instructions don't specify the exact word count requirement, models trained on these data may lack the ability to follow the writing instruction with word count requirements. Therefore, we sample 5,000 data from LongWrite-V-22k and calculate the length of the output $L$ then add a requirement "Please write L-word in total." to the end of the instruction and use gpt-4o-mini for rephrasing the instruction to maintain consistency. This is inspired by previous backtranslation~\cite{li2023self} method on training long-output LLMs~\cite{pham2024suri}.

\subsubsection{DPO Data Collection}
\label{sec:dpo_data_collection}

The SFT data aims to extend VLMs' output length. But the longer outputs may bring more hallucinations and repetitions. So the follow up question is: \textit{how to improve the generation fidelity of long output VLM?} Previous works often adapt direct preference optimization~\cite{rafailov2024direct,liu2024mia} to correct the hallucinations of VLMs. We follow the data format in RLHF-V~\cite{yu2024rlhf} which utilizes the human-annotated segment-level corrections on VLM's outputs as feedback.

\xhdr{VLM Output Collection} To collect long responses, we select 100 slides that were not included in LongWrite-V-22k for VLM to generate scripts. These slides were previously used for teaching on MOOC platforms~\cite{yu2020mooccube} and cover 10 subjects such as Computer Science, Math and Physics. Each subject may contain 4 to 16 slides and each slide may have 10 to 30 pages. We use LongWriter-V-7B, the VLM trained on our SFT data, to generate scripts for each slide. The long scripts are segmented by sections and aligned with each page of the given slide. We find that LongWriter-V-7B tends to output fewer sections than the number of total pages, which is one of the issues that we would ask human annotators to fix.

\xhdr{Human Revision Collection} To get high-quality feedback on the flawed output of SFT model, we hire 10 college students from 10 different majors corresponding to the subjects of our slides. We required annotators to have a GPA above 3.8 to ensure their expertise. To facilitate the annotation process, we build an online platform (See Appendix~\ref{sec:platform}). Each annotator will get slides that match with their major.The platform displays each slide page alongside the corresponding script segment generated by the SFT model. We ask annotators to check and revise each page's script for the following error types: factual errors, missing information, relevance to the image, coherence of sentences, and repetition of words. After completing the annotation of a slide, our authors will review the annotation quality. Ultimately, we get 72 valid scripts with fine-grained human corrections.

\begin{figure}[t]
    \centering
    \includegraphics[width=\linewidth]{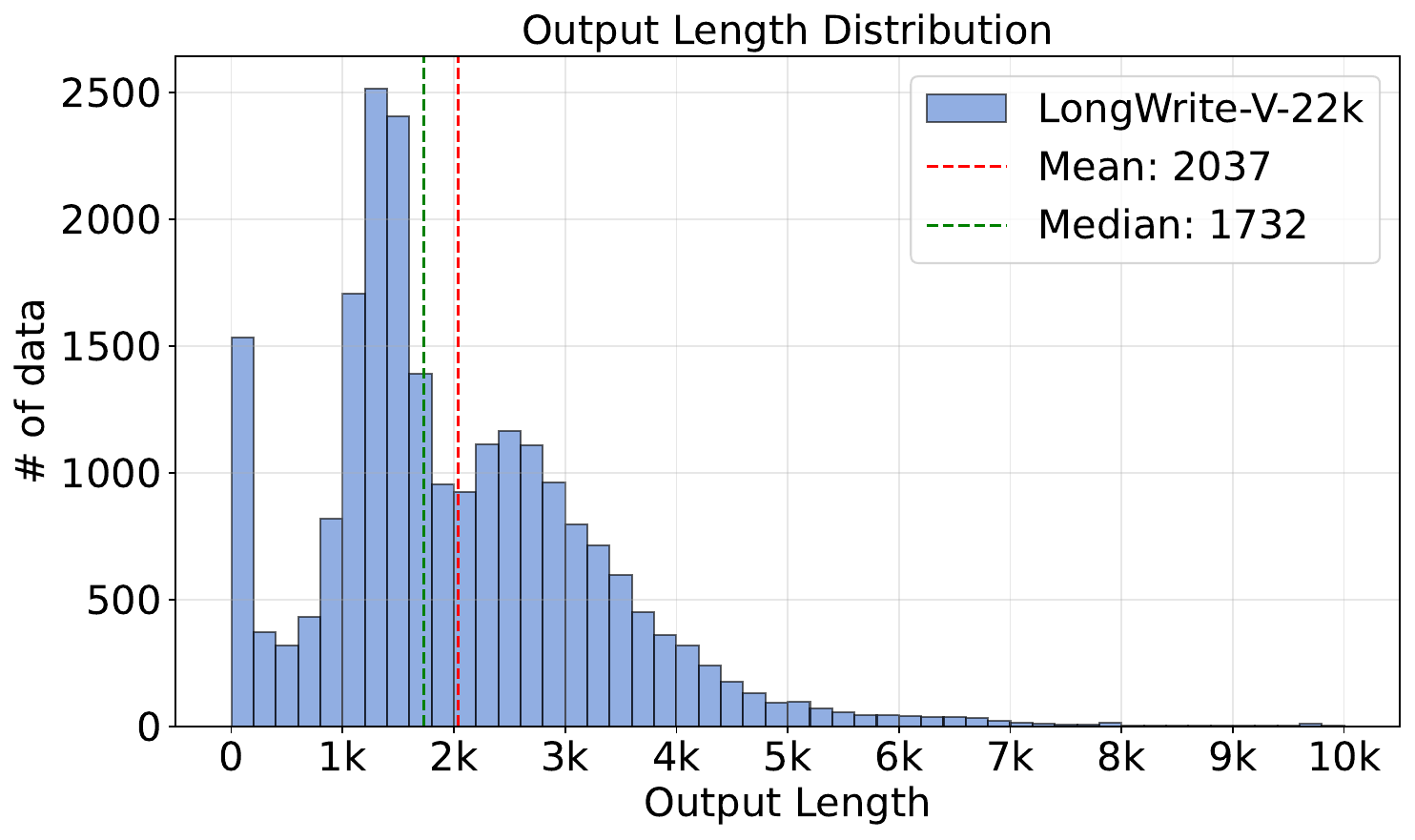}
    \caption{Output length statistics of LongWrite-V-22k.}
    \label{fig:stats}
\end{figure}

\subsection{Training}
\xhdr{Supervised Fine-tuning} We conduct model training based on two open-source VLMs with different parameter sizes: Qwen2.5-VL-7B-Instruct and Qwen2.5-VL-72B-Instruct~\cite{Qwen2.5-VL}. We choose Qwen2.5-VL series as base model because they support a context window of 32k tokens. By resizing the input image's width and height to 280x280, the Qwen2.5-VL models can process up to 30 images. As shown in Figure~\ref{fig:stats}, the output length in LongWrite-V-22k are distributed between 0 and 10k with two peaks around 0 and 1.5k. The peak at 0 indicates some short output data is mixed in the LongWrite-V-22k, which are mainly the results of those simple instructions. To get a better length distribution, we sample 10k data from LongWrite-V-22k with an average output length of 2.8k as training data. We then fine-tune the two models for 3 epochs with a learning rate of 1e-5 for Qwen2.5-VL-7B-Instruct and 7e-6 for Qwen2.5-VL-72B-Instruct, resulting in two SFT models: LongWriter-V-7B and LongWriter-V-72B.

\xhdr{Iterative Direct Preference Optimization} 
After SFT phase, DPO~\cite{rafailov2024direct} is a widely-used method to optimize VLM's output quality, which learns from a dataset of preference pairs $\mathcal{D}=\{(v,x, y_w, y_l)\}$, where the winning output $y_w$ is preferred over the losing output $y_l$ given the same visual input $v$ and text input $x$. The optimization objective of DPO is to maximize the difference between likelihood of preference pairs:  
\begin{equation}
\begin{aligned}
&\mathcal{L}_\text{DPO}(\pi_\theta; \pi_\text{ref})= -\mathbb{E}_{(v,x, y_w, y_l)\sim\mathcal{D}} \\
&[\log\sigma(\beta\log\frac{\pi_\theta(y_w|v,x)}{\pi_\text{ref}(y_w|v,x)}-\beta\log\frac{\pi_\theta(y_l|v,x)}{\pi_\text{ref}(y_l|v,x)})]
\end{aligned}
\end{equation}
In our annotation process, $v$ represents the images of a slide, $x$ is the instruction for generating scripts, $y_l$ is the flawed output script of VLM and $y_w$ is the slide's lecture after human revision. However, collecting human feedback on long output is very time-consuming and expensive. As mentioned in Section~\ref{sec:dpo_data_collection}, we gather 72 preference pairs on the scripts, which costs one week and around 1,000 \$ to finish.  To make most use of these data, we propose to iteratively learn the fine-grained human correctional feedback on the long output. As the $y_w = \{y_w^{i}, i=1,...N\}$ is a revised script for an $N$ page slide, we increasingly view each page's script $y_w^{i}$ as a winning segment over the flawed script:

\begin{small} 
\begin{equation}
\begin{aligned}
&\mathcal{L}_\text{IterDPO}(\pi_\theta; \pi_\text{ref})= -\mathbb{E}_{(v,x, y_w, y_l)\sim\mathcal{D}} \sum_{i=1}^{N}\\
&[\log\sigma(\beta\log\frac{\pi_\theta(y_w^{\le i}|v_{\le i},x)}{\pi_\text{ref}(y_w^{\le i}|v_{\le i},x)}-\beta\log\frac{\pi_\theta(y_l^{\le i}|v_{\le i},x)}{\pi_\text{ref}(y_l^{\le i}|v_{\le i},x)})]
\end{aligned}
\end{equation}
\end{small}
where $y_w^{\le i}$,  $y_l^{\le i}$ is the revised and unrevised scripts until page $i$, and $v_{\le i}$ are the corresponding images. We view $y_w^{\le i}$ as a new wining response over the flawed output $y_l^{\le i}$, this can help VLM learn the fine-grained feedback on the long output and extend the number preference pairs for $N$ times. In this way, we get 1,477 iterative pairs for training. Apart from human feedback, we also utilize AI feedback by employing the gpt4o as the reward model. Following RLAIF~\cite{yu2024rlaif}, we sample responses from the SFT model for 1,367 long-output instructions and use GPT-4o for assigning length and quality scores for the responses to construct preference pairs. Our final DPO model is trained with 2,844 mixed preference pairs,


\begin{table*}[t]
\centering
\resizebox{\linewidth}{!}{
\begin{tabular}{p{5.7cm}|m{0.7cm}m{0.7cm}m{0.7cm}|m{0.7cm}m{0.7cm}|m{0.7cm}m{0.7cm}|m{0.7cm}m{0.7cm}|m{0.7cm}m{0.7cm}}
\toprule
 &  \multicolumn{3}{|c}{\textbf{Overall}} &  \multicolumn{2}{|c}{\textbf{[0,1500)}} &  \multicolumn{2}{|c}{\textbf{[1500,2000)}} &  \multicolumn{2}{|c}{\textbf{[2000,3000)}} &  \multicolumn{2}{|c}{\textbf{[3000,4000)}} \\
\cmidrule(r){1-1} \cmidrule(lr){2-4} \cmidrule(lr){5-6} \cmidrule(lr){7-8} \cmidrule(lr){9-10} \cmidrule(lr){11-12}
 \textbf{Model} & $\overline{S}$ & $S_l$ & $S_q$ & $S_l$ & $S_q$ & $S_l$ & $S_q$ & $S_l$ & $S_q$ & $S_l$ & $S_q$ \\
\midrule
\multicolumn{12}{l}{\emph{Caption + LLMs}} \\
\texttt{GLM-4-9B-Chat} & 71.3 & 62.0 & 80.6 & 87.9 & 72.2 & 65.7 & 82.4 & 44.7 & 76.7 & 24.2 & 93.5 \\
\texttt{GPT-4o-2024-08-06} & 77.1 & 66.6 & 87.5 & 86.7 & 81.2 & 68.9 & 88.3 & 58.7 & 85.8 & 33.5 & 97.2 \\
\texttt{Mistral-Large-Instruct-2407} & 78.9 & 69.6 & 88.2 & \textbf{89.7} & 84.7 & 70.9 & 89.9 & 58.4 & 83.0 & 47.2 & 94.9 \\
\texttt{DeepSeek-R1} & 82.4 & 70.3 & \textbf{94.5} & 87.2 & 92.4 & 73.4 & \textbf{95.7} & 59.8 & \textbf{92.0} & 38.1 & \textbf{95.8} \\    
\midrule
\multicolumn{12}{l}{\emph{Open-source VLMs}} \\
\texttt{MiniCPM-V2.6} & 54.1 & 30.3 & 77.8 & 56.1 & 68.9 & 31.3 & 81.7 & 15.0 & 69.4 & 4.5 & 86.1 \\
\texttt{Qwen2.5-VL-7B-Instruct} & 54.4 & 45.3 & 63.5 & 62.9 & 51.1 & 46.6 & 70.5 & 37.6 & 50.6 & 16.1 & 67.6 \\
\texttt{Qwen2.5-VL-72B-Instruct} & 83.3 & 79.9 & 86.7 & 80.0 & 78.4 & 84.5 & 90.3 & 71.6 & 79.7 & 65.3 & 91.7 \\
\midrule
\multicolumn{12}{l}{\emph{Proprietary VLMs}} \\
\texttt{Claude-3-Opus-20240229} & 61.7 & 41.5 & 82.0 & 52.0 & 64.7 & 42.8 & 87.5 & 36.1 & 74.6 & 23.3 & 89.8 \\
\texttt{GPT-4o-2024-08-06} & 62.7 & 42.7 & 82.6 & 86.6 & 91.2 & 37.7 & 83.1 & 34.2 & 71.6 & 14.2 & 88.4 \\
\texttt{Gemini-1.5-Pro} & 83.0 & 74.8 & 91.2 & 88.7 & \textbf{93.0} & 78.1 & 91.8 & 62.2 & 86.2 & 50.5 & 95.4   \\

\midrule
\multicolumn{12}{l}{\emph{Our trained VLMs}} \\
\texttt{LongWriter-V-7B} & 81.8 & 82.5 & 81.1 & 63.3 & 72.8 & 87.8 & 86.4 & 81.2 & 69.2 & 86.8 & 87.5 \\
\texttt{LongWriter-V-7B-DPO} & 84.6 & \textbf{86.2} & 82.9 & 69.5 & 82.5 & \textbf{90.5} & 86.9 & 87.1 & 69.0 & \textbf{87.4} & 85.2 \\
\texttt{LongWriter-V-72B} & \textbf{84.9} & 84.3 & 85.5 & 73.2 & 83.3 & 86.2 & 89.3 & \textbf{88.4} & 75.8 & 81.4 & 85.2 \\
\bottomrule
\end{tabular}
}
\caption{Evaluation results (\%) on MMLongBench-Write. Note that LLMs are tested with input images transformed into captions. We report scores on different subsets of the benchmark, where [0,1000) means the expected output length falls within 0 to 1000 tokens. $\overline{S}$, $S_l$, $S_q$ is the overall score, length score and quality score respectively. }
\label{tb:exp}
\end{table*}

\section{Experiments}

\subsection{Experimental Setup}

\xhdr{Metric} Following~\citet{bai2024longwriter}, we evaluate the VLM's output length and quality using two metrics: $S_l$ and $S_q$. $S_l$ is the output score that measures how close that the VLM's output length $l_{v}$ is to the required length $l_{r}$:
\begin{equation}
S_l = 
\begin{cases} 
100 \cdot \max \left(0, 1 - \frac{(l_{v}/l_{r} - 1)}{3}\right) & \text{if } l_{v} > l_{r}, \\
100 \cdot \max \left(0, 1 - \frac{(l_{r}/l_{v} - 1)}{2}\right) & \text{if } l_{v} \leq l_{r}.
\end{cases}
\end{equation}
We also use gpt-4o-2024-08-06 to assign the quality score $S_q$ for six aspects: Relevance, Accuracy, Coherence, Clarity,
Breadth and Depth, and Reading Experience. We list the scoring prompt in Appendix~\ref{sec:setup}. Note that we have asked gpt-4o not to take the output length into account so that the quality score is independent with the length score. The overall score $\overline{S}$ is the mean of $S_l$ and $S_q$.

\xhdr{Baselines} We evaluate 3 proprietary VLMs, 3 open-source VLMs and 4 LLMs on MMLongBench-Write (model details about models are listed in Table~\ref{tb:model_card}). Given that LLMs can also process visual instructions via reading the image caption~\cite{ma2024mmlongbench}, we first use gpt-4o to describe the input images and then feed the caption and writing instruction to the LLM.

\subsection{Main Results}
We report the performance of baselines and our trained models in Table~\ref{tb:exp}. To study the effective output length of models, we divide the MMLongBench-Write benchmark into four subsets based on the instruction's required word count: 0-1500 words, 1500-2000 words, 2000-3000 words, and over 3000 words. The highest length and quality scores for each subset among models are in bold.
We have three observations on the results: (1) Most existing models struggle to satisfy the length requirement over 2000 words, while LongWriter-V models can generate enough words for such instructions. By checking the length score $S_l$ across different length intervals, we find that most models perform poorly on the [2000, 3000) range, with their $S_l$ below 70. In contrast, our LongWriter-V models can generate outputs with effective length and high quality even on the range of [3000, 4000).
(2) The scaling law effect on our benchmark is striking: smaller models like Qwen2.5-VL-7B-Instruct perform poorly in our evaluation with an overall score of 54.4, while its larger counterpart Qwen2.5-VL-72B-Instruct achieves a notably higher score of 83.3. Besides, after training the two VLMs on our LongWrite-V-22k data, both models improve significantly on long generation. The performance gap between the two sizes' models is narrowed after SFT (LongWriter-V-7B's 81.8 vs. LongWriter-V-72B's 84.9).
(3) DPO can improve both the VLM's output quality and the ability to follow the length requirements of long generation. LongWriter-V-7B-DPO, which is the model trained on LongWriter-V-7B with 2,844 preference pairs, achieves improvement on both $S_l$ (+3.7) and $S_q$ (+1.8),  showing that DPO is effective for boosting the long generation capabilities of VLMs.

\subsection{Human Evaluation}

As the quality score $S_q$ is assigned by the GPT-4o automatically, the evaluation results may have bias as LLM tends to favor the responses generated by itself~\cite{wang2023large,li2024llms}. To get a more fair quality comparison for the models, we conduct a human evaluation to capture the actual human preferences on model responses. Specifically, we select responses from four models: the three models trained by us and the GPT-4o-2024-08-06  baseline. We ask two human annotators to vote for their preferred response between two selected models on the $120$ responses of MMLongBench-Write. For each annotator, we collect $720$  votes and calculate the average win rate among models using two annotators' feedback. 

\begin{figure}[t]
    \centering
    \includegraphics[width=\linewidth]{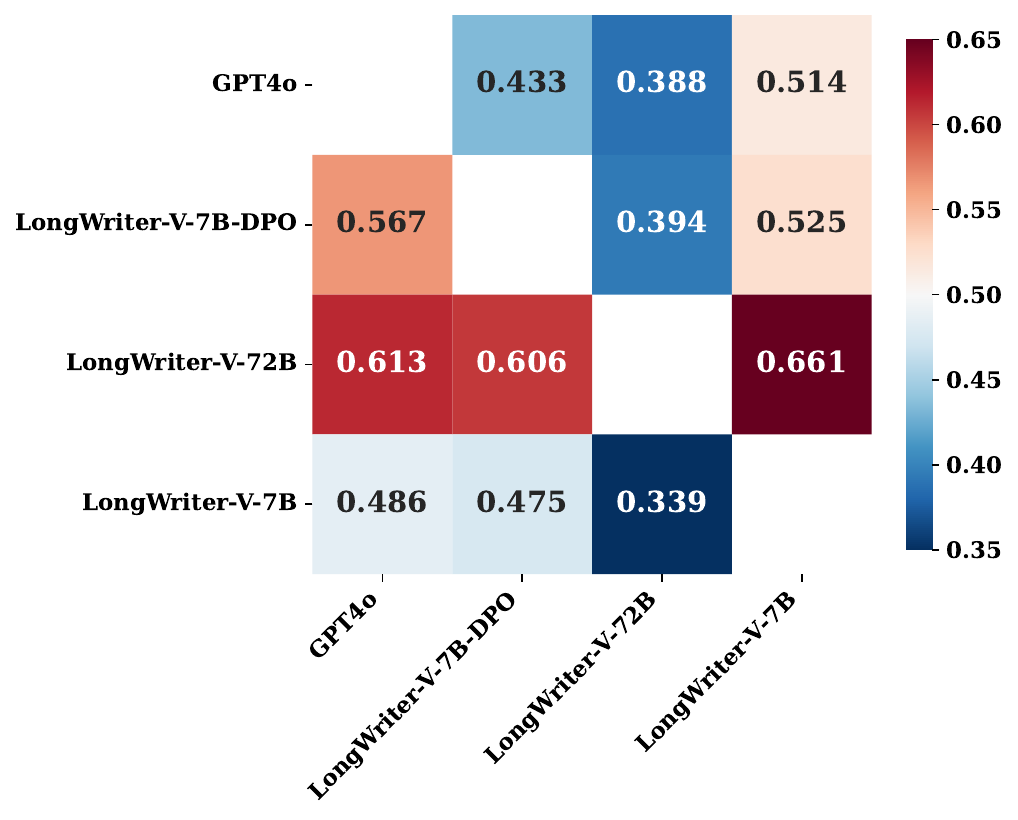}
    \caption{Human evaluation results on MMLongBench-Write, where each block of the matrix represents the model of the row's win rate over the model of the column. The win rate is voted by two annotators.}
    \label{fig:human_eval}
\end{figure}

The results are shown in Figure~\ref{fig:human_eval}, where we surprisingly find that two of our trained models receive more votes from humans in the comparison with the GPT-4o-2024-08-06 baseline. While in the automatic quality score comparison, the two models also surpass the GPT-4o on the quality score. This indicates that our trained models have gained some advantages over the GPT-4o baseline in the human preference, which is consistent with the automatic evaluation on the quality score of responses.

\subsection{Ablation Study}

We conduct ablation experiments on both the SFT and DPO process of LongWriter-V models. For the LongWriter-V-7B model trained on LongWrite-V-22k data, we control the three data sources of LongWrite-V-22k to observe how they contribute to the final performance of the SFT model.  We run the SFT process on Qwen2.5-VL-7B-Instruct without (w/o) single-image, multi-image or backtranslation data respectively and evaluate the trained models on MMLongBench-Write. As shown in Table~\ref{tb:mem}, removing any of these data sources may lead to a decline in the overall score, where multi-image data is the most essential one, causing a decrease of 15.3 overall score. These results indicate that these sources are useful for training long output VLMs.

To explore the effectiveness of our iterative DPO strategy over the small size preference data on long-output VLM alignment, we run the DPO process without those extra pairs extended by the iterative strategy. Results in Table~\ref{tb:mem} demonstrates that the model gains +1.1 length score but -1.1 quality score and -2.5 PPT task score over the DPO model with full data, which means the extended data is useful for the generation quality and the PPT script task. To examine the effectiveness of mixing AI preference pairs, we then train the SFT model with the human revised preference pairs only, resulting in a even worse performance (-1.1 overall score against the SFT model). This suggests that incorporating AI-generated pairs can improve model performance by providing additional training signals.

\section{Related Work}

Recent advancements in Vision-Language Models have focused on enhancing their ability to process long-context inputs~\cite{ge2024v2pe,li2024giraffe,chen2024internvl}. There are abundant benchmarks and datasets that designed for multimodal long context understanding including MMLongBench-Doc~\cite{ma2024mmlongbench}, LongDocURL~\cite{deng2024longdocurl}, LongViTU~\cite{wu2025longvitu}, ShareGPT4Video~\cite{chen2024sharegpt4video}, LongVideoBench ~\cite{wu2024longvideobench} and LVBench~\cite{wang2024lvbench}. However, the long-output generation abilities of VLMs have been less explored. In our work, we find that current VLMs struggle to generate an output with over 1000 tokens, which is much shorter than their max input context length (>16,000 tokens)~\cite{wang2024qwen2}. To fill this gap, we explore how to extend the maximum output length of VLMs.

Although we show that supervised fine-tuning can align VLMs with user's instructions on length requirements, it is also important to improve the quality of long output~\cite{wu2024longgenbench}. Previous works mainly focus on how to improve VLMs' generation quality on short output tasks via post training methods such as RLHF-V~\cite{yu2024rlhf}, RLAIF-V~\cite{yu2024rlaif}, POVID~\cite{zhou2024aligning} and MIA-DPO~\cite{liu2024mia}. However, none of these methods have explored how to effectively use human correctional feedback on long output for aligning VLMs. We propose to iteratively use each segment of the revised long output as the preferred response, which extends the number of preference pairs and successfully improves the long generation quality of VLM.

\begin{table}[t]
\centering
\resizebox{\linewidth}{!}{
\begin{tabular}{lcccc}
\toprule
\textbf{Model} & \textbf{$S$} & \textbf{$S_l$} & \textbf{$S_q$}   & \textbf{$S_{PPT}$} \\
\midrule
\texttt{LongWriter-V-7B} & 81.8 & 82.5 & 81.1 & 83.1 \\
\quad w/o single-image data & 79.6 & 79.5 & 79.6 & 83.4  \\
\quad w/o multi-image data & 66.5 & 60.3 & 72.7 & 29.3   \\
\quad w/o backtranslation & 80.7 & 80.0 & 81.3 & 82.4 \\
\midrule
\texttt{LongWriter-V-7B-DPO} & 84.6 & 86.3 & 82.9 & 85.8 \\
\quad w/o iterative pairs & 84.6 & 87.4 & 81.8 & 83.3  \\
\quad w/o 1.4k gpt4o feedback & 80.7 & 78.7 & 82.7 & 71.7 \\
\bottomrule
\end{tabular}
}
\caption{Scores (\%) on MMLongBench-Write for models trained under different conditions, where $S$, $S_l$ and $S_q$ is the overall, length and quality score on all tasks and $S_{PPT}$ is the overall score on the PPT script task.}
\label{tb:mem}
\end{table}
\section{Conclusion}

Our work introduces MMLongBench-Write, a comprehensive benchmark for evaluating long-generation tasks with visual inputs, and LongWriter-V-22k, a novel supervised fine-tuning dataset designed to enhance the long-output capabilities of VLMs. Furthermore, our proposed IterDPO method effectively leverages human feedback to improve the fidelity of long outputs, addressing issues such as hallucination. Future research may explore more efficient training strategies and larger datasets to further push the boundaries of long-output generation in VLMs.

\section*{Limitations}
We acknowledge some limitations in our work, which are listed below:
1. \textbf{Dataset Size}: The size of our LongWriter-V-22k dataset may not be sufficiently large to fully capture the diversity of long-output generation tasks. While this dataset size is adequate for initial exploration and training, it may limit the robustness of our findings and the generalizability of our model's performance. Expanding the dataset to include more examples would require significant additional resources, both in terms of data collection and annotation costs.
2. \textbf{Language Limitation}: The current dataset and benchmark are limited to English and Chinese only. This restricts our ability to evaluate the performance of VLMs across multiple languages, which is crucial for real-world applications where multilingual support is often required. Future work should consider expanding the dataset to include other languages to provide a more comprehensive evaluation of VLMs' long-output capabilities. 
3. \textbf{Human Feedback Efficiency}: While our IterDPO method significantly improves the efficiency of utilizing human feedback for long outputs, the process of collecting high-quality human corrections remains time-consuming and costly. This limits the scalability of our approach and the frequency with which we can update and refine the training data. Future work should explore more efficient methods for obtaining and incorporating human feedback to further enhance model performance.

\section*{Ethical Consideration}
While our work on enhancing the long-output capabilities of Vision-Language Models represents a significant advancement,  VLMs may still generate inaccurate or misleading information, especially when dealing with common knowledge not explicitly present in the context. This can lead to the propagation of false information if not properly managed. Therefore, additional safeguards and verification mechanisms should be implemented when deploying these models in user-facing applications.

Our training data has been desensitized to protect individual privacy. All the data sources we used are public available with permissible licenses, including MMEvol~\cite{luo2024mmevol} and Zenodo10K~\cite{zheng2025pptagent}.


\bibliography{custom}

\begin{thebibliography}{48}
\providecommand{\natexlab}[1]{#1}

\bibitem[{Alayrac et~al.(2022)Alayrac, Donahue, Luc, Miech, Barr, Hasson, Lenc, Mensch, Millican, Reynolds et~al.}]{alayrac2022flamingo}
Jean-Baptiste Alayrac, Jeff Donahue, Pauline Luc, Antoine Miech, Iain Barr, Yana Hasson, Karel Lenc, Arthur Mensch, Katherine Millican, Malcolm Reynolds, et~al. 2022.
\newblock Flamingo: a visual language model for few-shot learning.
\newblock \emph{Advances in neural information processing systems}, 35:23716--23736.

\bibitem[{Anthropic(2024)}]{claude-3}
Anthropic. 2024.
\newblock \href {https://www.anthropic.com/news/claude-3-family} {Anthropic: Introducing the next generation of claude}.

\bibitem[{Bai et~al.(2024)Bai, Zhang, Lv, Zheng, Zhu, Hou, Dong, Tang, and Li}]{bai2024longwriter}
Yushi Bai, Jiajie Zhang, Xin Lv, Linzhi Zheng, Siqi Zhu, Lei Hou, Yuxiao Dong, Jie Tang, and Juanzi Li. 2024.
\newblock Longwriter: Unleashing 10,000+ word generation from long context llms.
\newblock \emph{arXiv preprint arXiv:2408.07055}.

\bibitem[{Chen et~al.(2024{\natexlab{a}})Chen, Li, Dong, Zhang, He, Wang, Zhao, and Lin}]{chen2024sharegpt4v}
Lin Chen, Jinsong Li, Xiaoyi Dong, Pan Zhang, Conghui He, Jiaqi Wang, Feng Zhao, and Dahua Lin. 2024{\natexlab{a}}.
\newblock Sharegpt4v: Improving large multi-modal models with better captions.
\newblock In \emph{European Conference on Computer Vision}, pages 370--387. Springer.

\bibitem[{Chen et~al.(2024{\natexlab{b}})Chen, Wei, Li, Dong, Zhang, Zang, Chen, Duan, Lin, Tang et~al.}]{chen2024sharegpt4video}
Lin Chen, Xilin Wei, Jinsong Li, Xiaoyi Dong, Pan Zhang, Yuhang Zang, Zehui Chen, Haodong Duan, Bin Lin, Zhenyu Tang, et~al. 2024{\natexlab{b}}.
\newblock Sharegpt4video: Improving video understanding and generation with better captions.
\newblock \emph{arXiv preprint arXiv:2406.04325}.

\bibitem[{Chen et~al.(2024{\natexlab{c}})Chen, Wu, Wang, Su, Chen, Xing, Zhong, Zhang, Zhu, Lu et~al.}]{chen2024internvl}
Zhe Chen, Jiannan Wu, Wenhai Wang, Weijie Su, Guo Chen, Sen Xing, Muyan Zhong, Qinglong Zhang, Xizhou Zhu, Lewei Lu, et~al. 2024{\natexlab{c}}.
\newblock Internvl: Scaling up vision foundation models and aligning for generic visual-linguistic tasks.
\newblock In \emph{Proceedings of the IEEE/CVF Conference on Computer Vision and Pattern Recognition}, pages 24185--24198.

\bibitem[{Chou et~al.(2024)Chou, Dunlap, Mashita, Mandal, Darrell, Stoica, Gonzalez, and Chiang}]{chou2024visionarena}
Christopher Chou, Lisa Dunlap, Koki Mashita, Krishna Mandal, Trevor Darrell, Ion Stoica, Joseph~E Gonzalez, and Wei-Lin Chiang. 2024.
\newblock Visionarena: 230k real world user-vlm conversations with preference labels.
\newblock \emph{arXiv preprint arXiv:2412.08687}.

\bibitem[{Deng et~al.(2024)Deng, Yuan, Bu, Wang, Li, Xu, Li, Gao, Song, Zheng et~al.}]{deng2024longdocurl}
Chao Deng, Jiale Yuan, Pi~Bu, Peijie Wang, Zhong-Zhi Li, Jian Xu, Xiao-Hui Li, Yuan Gao, Jun Song, Bo~Zheng, et~al. 2024.
\newblock Longdocurl: a comprehensive multimodal long document benchmark integrating understanding, reasoning, and locating.
\newblock \emph{arXiv preprint arXiv:2412.18424}.

\bibitem[{Favero et~al.(2024)Favero, Zancato, Trager, Choudhary, Perera, Achille, Swaminathan, and Soatto}]{favero2024multi}
Alessandro Favero, Luca Zancato, Matthew Trager, Siddharth Choudhary, Pramuditha Perera, Alessandro Achille, Ashwin Swaminathan, and Stefano Soatto. 2024.
\newblock Multi-modal hallucination control by visual information grounding.
\newblock In \emph{Proceedings of the IEEE/CVF Conference on Computer Vision and Pattern Recognition}, pages 14303--14312.

\bibitem[{Ge et~al.(2024)Ge, Chen, Lin, Zhu, Liu, Dai, and Zhu}]{ge2024v2pe}
Junqi Ge, Ziyi Chen, Jintao Lin, Jinguo Zhu, Xihui Liu, Jifeng Dai, and Xizhou Zhu. 2024.
\newblock V2pe: Improving multimodal long-context capability of vision-language models with variable visual position encoding.
\newblock \emph{arXiv preprint arXiv:2412.09616}.

\bibitem[{GLM et~al.(2024)GLM, Zeng, Xu, Wang, Zhang, Yin, Rojas, Feng, Zhao, Lai, Yu, Wang, Sun, Zhang, Cheng, Gui, Tang, Zhang, Li, Zhao, Wu, Zhong, Liu, Huang, Zhang, Zheng, Lu, Duan, Zhang, Cao, Yang, Tam, Zhao, Liu, Xia, Zhang, Gu, Lv, Liu, Liu, Yang, Song, Zhang, An, Xu, Niu, Yang, Li, Bai, Dong, Qi, Wang, Yang, Du, Hou, and Wang}]{glm2024chatglm}
Team GLM, Aohan Zeng, Bin Xu, Bowen Wang, Chenhui Zhang, Da~Yin, Diego Rojas, Guanyu Feng, Hanlin Zhao, Hanyu Lai, Hao Yu, Hongning Wang, Jiadai Sun, Jiajie Zhang, Jiale Cheng, Jiayi Gui, Jie Tang, Jing Zhang, Juanzi Li, Lei Zhao, Lindong Wu, Lucen Zhong, Mingdao Liu, Minlie Huang, Peng Zhang, Qinkai Zheng, Rui Lu, Shuaiqi Duan, Shudan Zhang, Shulin Cao, Shuxun Yang, Weng~Lam Tam, Wenyi Zhao, Xiao Liu, Xiao Xia, Xiaohan Zhang, Xiaotao Gu, Xin Lv, Xinghan Liu, Xinyi Liu, Xinyue Yang, Xixuan Song, Xunkai Zhang, Yifan An, Yifan Xu, Yilin Niu, Yuantao Yang, Yueyan Li, Yushi Bai, Yuxiao Dong, Zehan Qi, Zhaoyu Wang, Zhen Yang, Zhengxiao Du, Zhenyu Hou, and Zihan Wang. 2024.
\newblock Chatglm: A family of large language models from glm-130b to glm-4 all tools.
\newblock \emph{arXiv preprint arXiv:2406.12793}.

\bibitem[{Guo et~al.(2025)Guo, Yang, Zhang, Song, Zhang, Xu, Zhu, Ma, Wang, Bi et~al.}]{guo2025deepseek}
Daya Guo, Dejian Yang, Haowei Zhang, Junxiao Song, Ruoyu Zhang, Runxin Xu, Qihao Zhu, Shirong Ma, Peiyi Wang, Xiao Bi, et~al. 2025.
\newblock Deepseek-r1: Incentivizing reasoning capability in llms via reinforcement learning.
\newblock \emph{arXiv preprint arXiv:2501.12948}.

\bibitem[{Hartsock and Rasool(2024)}]{hartsock2024vision}
Iryna Hartsock and Ghulam Rasool. 2024.
\newblock Vision-language models for medical report generation and visual question answering: A review.
\newblock \emph{Frontiers in Artificial Intelligence}, 7:1430984.

\bibitem[{Hong et~al.(2023)Hong, Sayeed, Mehra, Demberg, and Schiele}]{hong2023visual}
Xudong Hong, Asad Sayeed, Khushboo Mehra, Vera Demberg, and Bernt Schiele. 2023.
\newblock Visual writing prompts: Character-grounded story generation with curated image sequences.
\newblock \emph{Transactions of the Association for Computational Linguistics}, 11:565--581.

\bibitem[{Jiang et~al.(2023)Jiang, Sablayrolles, Mensch, Bamford, Chaplot, Casas, Bressand, Lengyel, Lample, Saulnier et~al.}]{jiang2023mistral}
Albert~Q Jiang, Alexandre Sablayrolles, Arthur Mensch, Chris Bamford, Devendra~Singh Chaplot, Diego de~las Casas, Florian Bressand, Gianna Lengyel, Guillaume Lample, Lucile Saulnier, et~al. 2023.
\newblock Mistral 7b.
\newblock \emph{arXiv preprint arXiv:2310.06825}.

\bibitem[{Li et~al.(2024{\natexlab{a}})Li, Dong, Chen, Su, Zhou, Ai, Ye, and Liu}]{li2024llms}
Haitao Li, Qian Dong, Junjie Chen, Huixue Su, Yujia Zhou, Qingyao Ai, Ziyi Ye, and Yiqun Liu. 2024{\natexlab{a}}.
\newblock Llms-as-judges: a comprehensive survey on llm-based evaluation methods.
\newblock \emph{arXiv preprint arXiv:2412.05579}.

\bibitem[{Li et~al.(2024{\natexlab{b}})Li, Li, Gong, and Liu}]{li2024giraffe}
Mukai Li, Lei Li, Shansan Gong, and Qi~Liu. 2024{\natexlab{b}}.
\newblock Giraffe: Design choices for extending the context length of visual language models.
\newblock \emph{arXiv preprint arXiv:2412.12735}.

\bibitem[{Li et~al.(2023)Li, Yu, Zhou, Schick, Levy, Zettlemoyer, Weston, and Lewis}]{li2023self}
Xian Li, Ping Yu, Chunting Zhou, Timo Schick, Omer Levy, Luke Zettlemoyer, Jason Weston, and Mike Lewis. 2023.
\newblock Self-alignment with instruction backtranslation.
\newblock \emph{arXiv preprint arXiv:2308.06259}.

\bibitem[{Lin et~al.(2014)Lin, Maire, Belongie, Hays, Perona, Ramanan, Doll{\'a}r, and Zitnick}]{lin2014microsoft}
Tsung-Yi Lin, Michael Maire, Serge Belongie, James Hays, Pietro Perona, Deva Ramanan, Piotr Doll{\'a}r, and C~Lawrence Zitnick. 2014.
\newblock Microsoft coco: Common objects in context.
\newblock In \emph{Computer Vision--ECCV 2014: 13th European Conference, Zurich, Switzerland, September 6-12, 2014, Proceedings, Part V 13}, pages 740--755. Springer.

\bibitem[{Liu et~al.(2024{\natexlab{a}})Liu, Li, Wu, and Lee}]{liu2024visual}
Haotian Liu, Chunyuan Li, Qingyang Wu, and Yong~Jae Lee. 2024{\natexlab{a}}.
\newblock Visual instruction tuning.
\newblock \emph{Advances in neural information processing systems}, 36.

\bibitem[{Liu et~al.(2024{\natexlab{b}})Liu, Zeng, Ren, Li, Zhang, Yang, Jiang, Li, Yang, Su et~al.}]{liu2024grounding}
Shilong Liu, Zhaoyang Zeng, Tianhe Ren, Feng Li, Hao Zhang, Jie Yang, Qing Jiang, Chunyuan Li, Jianwei Yang, Hang Su, et~al. 2024{\natexlab{b}}.
\newblock Grounding dino: Marrying dino with grounded pre-training for open-set object detection.
\newblock In \emph{European Conference on Computer Vision}, pages 38--55. Springer.

\bibitem[{Liu et~al.(2024{\natexlab{c}})Liu, Zang, Dong, Zhang, Cao, Duan, He, Xiong, Lin, and Wang}]{liu2024mia}
Ziyu Liu, Yuhang Zang, Xiaoyi Dong, Pan Zhang, Yuhang Cao, Haodong Duan, Conghui He, Yuanjun Xiong, Dahua Lin, and Jiaqi Wang. 2024{\natexlab{c}}.
\newblock Mia-dpo: Multi-image augmented direct preference optimization for large vision-language models.
\newblock \emph{arXiv preprint arXiv:2410.17637}.

\bibitem[{Luo et~al.(2024)Luo, Zhang, Chen, Lin, Liu, Wu, Yang, Wang, Zeng, Gao et~al.}]{luo2024mmevol}
Run Luo, Haonan Zhang, Longze Chen, Ting-En Lin, Xiong Liu, Yuchuan Wu, Min Yang, Minzheng Wang, Pengpeng Zeng, Lianli Gao, et~al. 2024.
\newblock Mmevol: Empowering multimodal large language models with evol-instruct.
\newblock \emph{arXiv preprint arXiv:2409.05840}.

\bibitem[{Ma et~al.(2024)Ma, Zang, Chen, Chen, Jiao, Li, Lu, Liu, Ma, Dong et~al.}]{ma2024mmlongbench}
Yubo Ma, Yuhang Zang, Liangyu Chen, Meiqi Chen, Yizhu Jiao, Xinze Li, Xinyuan Lu, Ziyu Liu, Yan Ma, Xiaoyi Dong, et~al. 2024.
\newblock Mmlongbench-doc: Benchmarking long-context document understanding with visualizations.
\newblock \emph{arXiv preprint arXiv:2407.01523}.

\bibitem[{OpenAI(2024)}]{GPT-4o}
OpenAI. 2024.
\newblock \href {https://openai.com/index/hello-gpt-4o/} {Openai: Hello gpt-4o}.

\bibitem[{Pham et~al.(2024)Pham, Sun, and Iyyer}]{pham2024suri}
Chau~Minh Pham, Simeng Sun, and Mohit Iyyer. 2024.
\newblock Suri: Multi-constraint instruction following for long-form text generation.
\newblock \emph{arXiv preprint arXiv:2406.19371}.

\bibitem[{Rafailov et~al.(2024)Rafailov, Sharma, Mitchell, Manning, Ermon, and Finn}]{rafailov2024direct}
Rafael Rafailov, Archit Sharma, Eric Mitchell, Christopher~D Manning, Stefano Ermon, and Chelsea Finn. 2024.
\newblock Direct preference optimization: Your language model is secretly a reward model.
\newblock \emph{Advances in Neural Information Processing Systems}, 36.

\bibitem[{Schuhmann et~al.(2022)Schuhmann, Beaumont, Vencu, Gordon, Wightman, Cherti, Coombes, Katta, Mullis, Wortsman et~al.}]{schuhmann2022laion}
Christoph Schuhmann, Romain Beaumont, Richard Vencu, Cade Gordon, Ross Wightman, Mehdi Cherti, Theo Coombes, Aarush Katta, Clayton Mullis, Mitchell Wortsman, et~al. 2022.
\newblock Laion-5b: An open large-scale dataset for training next generation image-text models.
\newblock \emph{Advances in Neural Information Processing Systems}, 35:25278--25294.

\bibitem[{Shu et~al.(2024)Shu, Zhang, Liu, Qin, Zhou, Huang, and Zhao}]{shu2024video}
Yan Shu, Peitian Zhang, Zheng Liu, Minghao Qin, Junjie Zhou, Tiejun Huang, and Bo~Zhao. 2024.
\newblock Video-xl: Extra-long vision language model for hour-scale video understanding.
\newblock \emph{arXiv preprint arXiv:2409.14485}.

\bibitem[{Taavitsainen and Pahta(2000)}]{taavitsainen2000conventions}
Irma Taavitsainen and P{\"a}ivi Pahta. 2000.
\newblock Conventions of professional writing: The medical case report in a historical perspective.
\newblock \emph{Journal of English Linguistics}, 28(1):60--76.

\bibitem[{Team et~al.(2024)Team, Georgiev, Lei, Burnell, Bai, Gulati, Tanzer, Vincent, Pan, Wang et~al.}]{team2024gemini}
Gemini Team, Petko Georgiev, Ving~Ian Lei, Ryan Burnell, Libin Bai, Anmol Gulati, Garrett Tanzer, Damien Vincent, Zhufeng Pan, Shibo Wang, et~al. 2024.
\newblock Gemini 1.5: Unlocking multimodal understanding across millions of tokens of context.
\newblock \emph{arXiv preprint arXiv:2403.05530}.

\bibitem[{Team(2025)}]{Qwen2.5-VL}
Qwen Team. 2025.
\newblock \href {https://qwenlm.github.io/blog/qwen2.5-vl/} {Qwen2.5-vl}.

\bibitem[{Wang et~al.(2022)Wang, Yang, Hu, Li, Lin, Gan, Liu, Liu, and Wang}]{wang2022git}
Jianfeng Wang, Zhengyuan Yang, Xiaowei Hu, Linjie Li, Kevin Lin, Zhe Gan, Zicheng Liu, Ce~Liu, and Lijuan Wang. 2022.
\newblock Git: A generative image-to-text transformer for vision and language.
\newblock \emph{arXiv preprint arXiv:2205.14100}.

\bibitem[{Wang et~al.(2023)Wang, Li, Chen, Cai, Zhu, Lin, Cao, Liu, Liu, and Sui}]{wang2023large}
Peiyi Wang, Lei Li, Liang Chen, Zefan Cai, Dawei Zhu, Binghuai Lin, Yunbo Cao, Qi~Liu, Tianyu Liu, and Zhifang Sui. 2023.
\newblock Large language models are not fair evaluators.
\newblock \emph{arXiv preprint arXiv:2305.17926}.

\bibitem[{Wang et~al.(2024{\natexlab{a}})Wang, Bai, Tan, Wang, Fan, Bai, Chen, Liu, Wang, Ge et~al.}]{wang2024qwen2}
Peng Wang, Shuai Bai, Sinan Tan, Shijie Wang, Zhihao Fan, Jinze Bai, Keqin Chen, Xuejing Liu, Jialin Wang, Wenbin Ge, et~al. 2024{\natexlab{a}}.
\newblock Qwen2-vl: Enhancing vision-language model's perception of the world at any resolution.
\newblock \emph{arXiv preprint arXiv:2409.12191}.

\bibitem[{Wang et~al.(2024{\natexlab{b}})Wang, Chen, Jia, Wang, Fang, Wang, Gao, Xie, Xu, Dai et~al.}]{wang2024weaver}
Tiannan Wang, Jiamin Chen, Qingrui Jia, Shuai Wang, Ruoyu Fang, Huilin Wang, Zhaowei Gao, Chunzhao Xie, Chuou Xu, Jihong Dai, et~al. 2024{\natexlab{b}}.
\newblock Weaver: Foundation models for creative writing.
\newblock \emph{arXiv preprint arXiv:2401.17268}.

\bibitem[{Wang et~al.(2024{\natexlab{c}})Wang, He, Hong, Cheng, Zhang, Qi, Gu, Huang, Xu, Dong et~al.}]{wang2024lvbench}
Weihan Wang, Zehai He, Wenyi Hong, Yean Cheng, Xiaohan Zhang, Ji~Qi, Xiaotao Gu, Shiyu Huang, Bin Xu, Yuxiao Dong, et~al. 2024{\natexlab{c}}.
\newblock Lvbench: An extreme long video understanding benchmark.
\newblock \emph{arXiv preprint arXiv:2406.08035}.

\bibitem[{Wu et~al.(2024{\natexlab{a}})Wu, Li, Chen, and Li}]{wu2024longvideobench}
Haoning Wu, Dongxu Li, Bei Chen, and Junnan Li. 2024{\natexlab{a}}.
\newblock Longvideobench: A benchmark for long-context interleaved video-language understanding.
\newblock \emph{arXiv preprint arXiv:2407.15754}.

\bibitem[{Wu et~al.(2025)Wu, Ma, Ci, Fan, Wang, Zhao, Li, and Wang}]{wu2025longvitu}
Rujie Wu, Xiaojian Ma, Hai Ci, Yue Fan, Yuxuan Wang, Haozhe Zhao, Qing Li, and Yizhou Wang. 2025.
\newblock Longvitu: Instruction tuning for long-form video understanding.
\newblock \emph{arXiv preprint arXiv:2501.05037}.

\bibitem[{Wu et~al.(2024{\natexlab{b}})Wu, Hee, Hu, and Lee}]{wu2024longgenbench}
Yuhao Wu, Ming~Shan Hee, Zhiqing Hu, and Roy Ka-Wei Lee. 2024{\natexlab{b}}.
\newblock Longgenbench: Benchmarking long-form generation in long context llms.
\newblock \emph{arXiv preprint arXiv:2409.02076}.

\bibitem[{Xue et~al.(2024)Xue, Chen, Li, Hu, Zhu, Li, Fang, Tang, Yang, Liu et~al.}]{xue2024longvila}
Fuzhao Xue, Yukang Chen, Dacheng Li, Qinghao Hu, Ligeng Zhu, Xiuyu Li, Yunhao Fang, Haotian Tang, Shang Yang, Zhijian Liu, et~al. 2024.
\newblock Longvila: Scaling long-context visual language models for long videos.
\newblock \emph{arXiv preprint arXiv:2408.10188}.

\bibitem[{Yao et~al.(2024)Yao, Yu, Zhang, Wang, Cui, Zhu, Cai, Li, Zhao, He et~al.}]{yao2024minicpm}
Yuan Yao, Tianyu Yu, Ao~Zhang, Chongyi Wang, Junbo Cui, Hongji Zhu, Tianchi Cai, Haoyu Li, Weilin Zhao, Zhihui He, et~al. 2024.
\newblock Minicpm-v: A gpt-4v level mllm on your phone.
\newblock \emph{arXiv preprint arXiv:2408.01800}.

\bibitem[{Yu et~al.(2020)Yu, Luo, Xiao, Zhong, Wang, Feng, Luo, Wang, Hou, Li et~al.}]{yu2020mooccube}
Jifan Yu, Gan Luo, Tong Xiao, Qingyang Zhong, Yuquan Wang, Wenzheng Feng, Junyi Luo, Chenyu Wang, Lei Hou, Juanzi Li, et~al. 2020.
\newblock Mooccube: A large-scale data repository for nlp applications in moocs.
\newblock In \emph{Proceedings of the 58th annual meeting of the association for computational linguistics}, pages 3135--3142.

\bibitem[{Yu et~al.(2024{\natexlab{a}})Yu, Yao, Zhang, He, Han, Cui, Hu, Liu, Zheng, Sun et~al.}]{yu2024rlhf}
Tianyu Yu, Yuan Yao, Haoye Zhang, Taiwen He, Yifeng Han, Ganqu Cui, Jinyi Hu, Zhiyuan Liu, Hai-Tao Zheng, Maosong Sun, et~al. 2024{\natexlab{a}}.
\newblock Rlhf-v: Towards trustworthy mllms via behavior alignment from fine-grained correctional human feedback.
\newblock In \emph{Proceedings of the IEEE/CVF Conference on Computer Vision and Pattern Recognition}, pages 13807--13816.

\bibitem[{Yu et~al.(2024{\natexlab{b}})Yu, Zhang, Yao, Dang, Chen, Lu, Cui, He, Liu, Chua et~al.}]{yu2024rlaif}
Tianyu Yu, Haoye Zhang, Yuan Yao, Yunkai Dang, Da~Chen, Xiaoman Lu, Ganqu Cui, Taiwen He, Zhiyuan Liu, Tat-Seng Chua, et~al. 2024{\natexlab{b}}.
\newblock Rlaif-v: Aligning mllms through open-source ai feedback for super gpt-4v trustworthiness.
\newblock \emph{arXiv preprint arXiv:2405.17220}.

\bibitem[{Zhang et~al.(2024)Zhang, Huang, Jin, and Lu}]{zhang2024vision}
Jingyi Zhang, Jiaxing Huang, Sheng Jin, and Shijian Lu. 2024.
\newblock Vision-language models for vision tasks: A survey.
\newblock \emph{IEEE Transactions on Pattern Analysis and Machine Intelligence}.

\bibitem[{Zheng et~al.(2025)Zheng, Guan, Kong, Zheng, Lin, Lu, He, Han, and Sun}]{zheng2025pptagent}
Hao Zheng, Xinyan Guan, Hao Kong, Jia Zheng, Hongyu Lin, Yaojie Lu, Ben He, Xianpei Han, and Le~Sun. 2025.
\newblock \href {https://arxiv.org/abs/2501.03936} {Pptagent: Generating and evaluating presentations beyond text-to-slides}.
\newblock \emph{Preprint}, arXiv:2501.03936.

\bibitem[{Zhou et~al.(2024)Zhou, Cui, Rafailov, Finn, and Yao}]{zhou2024aligning}
Yiyang Zhou, Chenhang Cui, Rafael Rafailov, Chelsea Finn, and Huaxiu Yao. 2024.
\newblock Aligning modalities in vision large language models via preference fine-tuning.
\newblock \emph{arXiv preprint arXiv:2402.11411}.

\end{thebibliography}

\newpage
\appendix
\onecolumn

\section*{Appendix}


\section{Model Cards}
Table~\ref{tb:model_card} demonstrates the detailed information of the LLMs and VLMs evaluated in our experiments.

\begin{table}[htbp]
    \centering
    \resizebox{\linewidth}{!}{
    \begin{tabular}{llrr}
    \toprule
    \textbf{Model name} & \textbf{Model version} & \textbf{Context window} & \textbf{Max output tokens} \\
    \midrule
    \emph{Large Language Models}  & \\
    GLM-4-9B-chat~\citep{glm2024chatglm} & - & 128,000 tokens & - \\
    Mistral-Large-Instruct~\citep{jiang2023mistral} & Mistral-Large-Instruct-2407 & 128,000 tokens & - \\
    Deepseek-r1~\cite{guo2025deepseek} & deepseek-reasoner & 64,000 tokens &  8,000 tokens  \\
    \midrule
    \emph{Vison Language Models}  & \\
    MiniCPM-V2.6~\citep{yao2024minicpm} & MiniCPM-V-2-6 & 32,000 tokens & - \\
     Qwen2.5-VL-7B~\citep{Qwen2.5-VL} &  Qwen2.5-VL-7B-Instruct & 32,000 tokens & - \\
    Qwen2.5-VL-72B~\citep{Qwen2.5-VL} &  Qwen2.5-VL-72B-Instruct &  32,000 tokens & - \\
    Claude 3 Opus~\citep{claude-3} & claude-3-opus-20240229 & 200,000 tokens & 4,096 tokens \\
    Gemini-1.5-pro~\citep{team2024gemini} & gemini-1.5-pro & 2,000,000 tokens & 8,192 tokens	 \\
    GPT-4o~\citep{GPT-4o} & gpt-4o-2024-08-06 & 128,000 tokens & 8,192 tokens \\
    
    \bottomrule
    \end{tabular}
    }
    \caption{Model cards.}
    \label{tb:model_card}
\end{table}

\section{Model Prompts}
\label{sec:task}

\subsection{Prompts for Collecting Visual Instructions}
\xhdr{Prompt for selecting user requests that require 1,000+ word response}
\begin{tcolorbox}[size=title,opacityfill=0.1,breakable]
\noindent
You will receive an image and an instruction from a user to an AI assistant, please determine whether the instruction requires the AI assistant to write an article for the given image, and the length of the article is more than 1,000 words in English (or 1,000 characters in Chinese). If the instruction does not mention the word requirement, please determine whether the user’s intention of the response length is more than 1,000 words. If the instruction is irrelated with the image, please reply “no”.
Instruction: \{\textit{User Instruction}\}
\end{tcolorbox}

\xhdr{Prompt for constructing multi-image instruction}
\begin{tcolorbox}[size=title,opacityfill=0.1,breakable]
\noindent
You will receive {\{\textit{Image Number}\}} images and an instruction from a user to an AI assistant, this original instruction is targeted for the first image solely. Now please rewrite this instruction to a challenging long-output one that need using visual information from all the input images, and the length of the expected output should be more than 2,000 words in English (or 2,000 characters in Chinese). Here are three examples of challenging long-output instructions:

Example instruction 1: \{\textit{Example Instruction 1}\}

Example instruction 2: \{\textit{Example Instruction 2}\}

Example instruction 3: \{\textit{Example Instruction 3}\}

Now, you should rewrite the following instruction:

Instruction: \{\textit{User Instruction }\}

Please rewrite this user instruction to a challenging long-output instruction that requires the use of all the input images. Please output only the rewritten instruction, do not output other content.
\end{tcolorbox}

\subsection{Prompts for the LongWrite Agent-V Pipeline}
\label{app:prompt_lav}

\xhdr{Prompt for planning the writing outline}
\begin{tcolorbox}[size=title,opacityfill=0.1,breakable]
\noindent
You are an expert planner. Your task is to break down a writing task into clear subtasks based on the provided images and writing instruction.

Please analyze the images and writing instruction carefully, then create a detailed outline in this format:

Section 1 - Main Point: [Key points to cover based on images and instruction] - Word Count: [200-1000 words]

Section 2 - Main Point: [Key points to cover based on images and instruction] - Word Count: [200-1000 words]

...

Make each section focused and specific while ensuring the full outline:

1. Covers all key content from both images and writing instruction

2. Flows logically from section to section

3. Has reasonable word count targets (200-1000 words per section)

4. Forms a cohesive whole that fulfills the writing instruction

Writing instruction: \{\textit{User Instruction}\}

Output only the outline with no other text.
\end{tcolorbox}

\xhdr{Prompt for generating each paragraph according to the writing outline}
\begin{tcolorbox}[size=title,opacityfill=0.1,breakable]
\noindent
You are an expert writer. Your task is to write the next section of a longer piece based on:

1. The provided images and writing instruction

2. The outline plan

3. Previously written sections

Writing instruction:  \{\textit{User Instruction}\}

Outline plan: \{\textit{PLAN}\}

Previous sections: \{\textit{TEXT}\}

Please write section  \{\textit{STEP}\} following these guidelines:

1. Focus on the main points specified in the outline

2. Stay within the target word count

3. Flow naturally from previous sections

4. Integrate relevant details from the images

5. Maintain a consistent tone and style

6. Write only this section, not a full conclusion

Output only the new section with no other text.
\end{tcolorbox}

\section{Annotation Details}

\subsection{Annotation Platform}
\label{sec:platform}

Our annotation platform consists of two main pages: annotation page and admin page.

\begin{figure}[t]
    \centering
    \includegraphics[width=\linewidth]{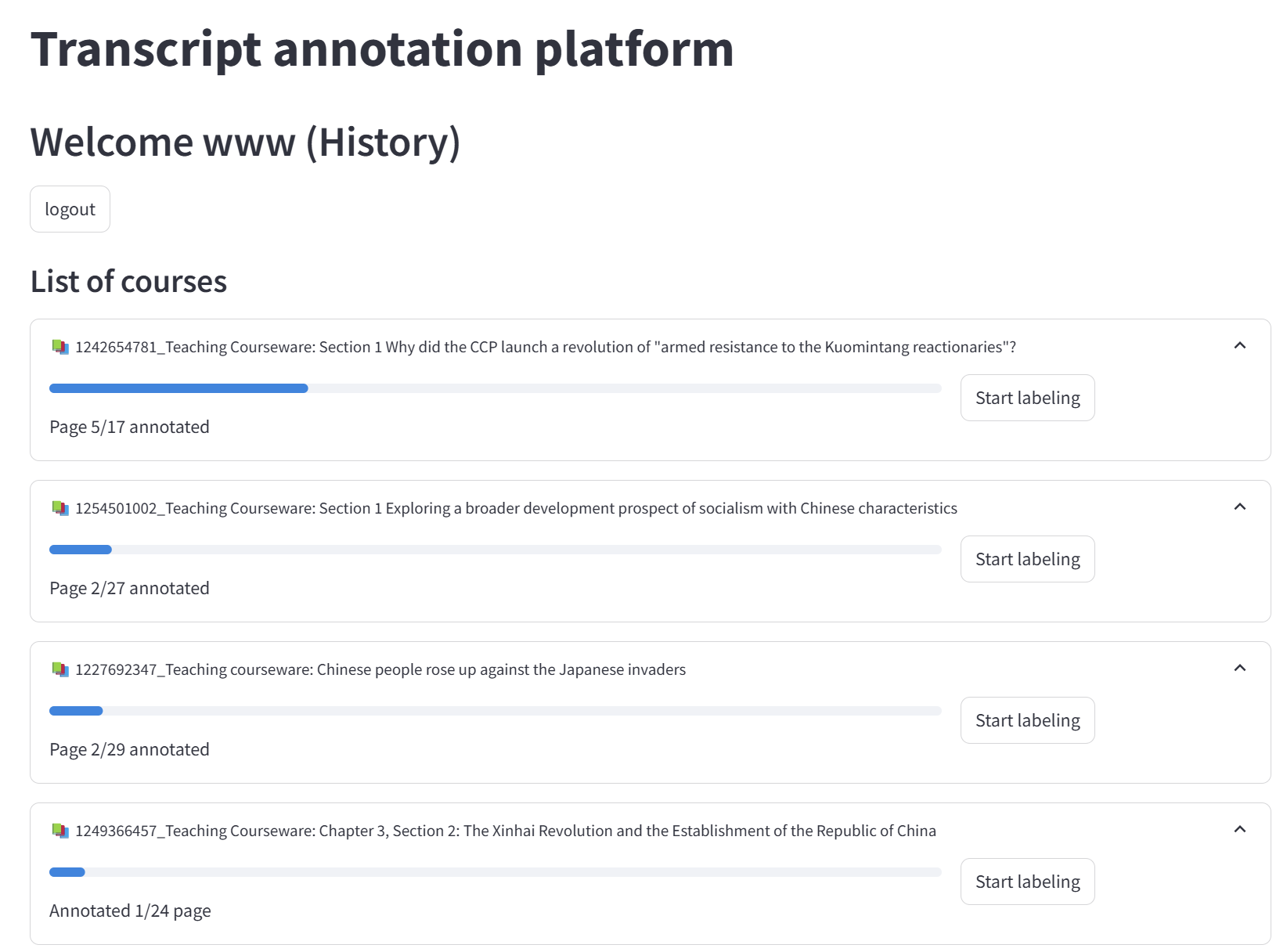}
    \caption{Screenshot of the \rednote{annotation page} (course selection). After logging in, annotators can view available courses in their major, with progress bars showing completion status and the number of annotated pages.}
    \label{fig:Annotation_Annotation1}
\end{figure}

\begin{figure*}[htbp]
    \centering
    \includegraphics[width=\linewidth]{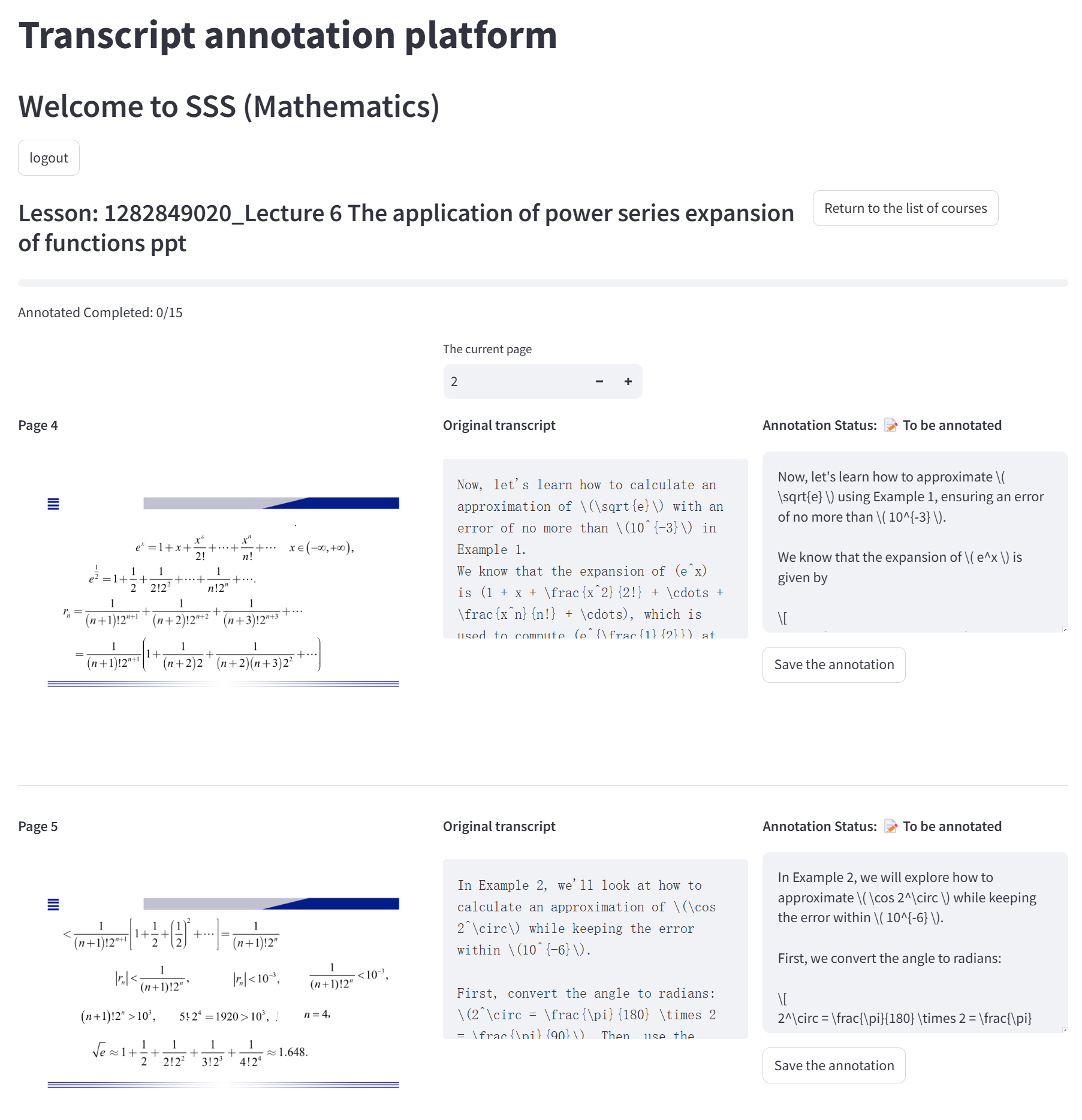}
    \caption{Screenshot of the \rednote{annotation page} (annotation). The page displays the lecture slide (left), original transcript (middle), and annotation area (right). Users can modify transcripts while viewing the corresponding slides and track their annotation status.}
    \label{fig:Annotation_Annotation2}
\end{figure*}

\noindent
\textbf{\rednote{Annotation page.}} 
This page provides the core annotation interface for users. As shown in Figure~\ref{fig:Annotation_Annotation1}, after logging in, users first see a course selection interface where they can view available courses in their major. Each course is displayed with a progress bar showing completion status and the total number of annotated pages. After selecting a course, users enter the annotation interface shown in Figure~\ref{fig:Annotation_Annotation2}, where the page is divided into three columns: the lecture slide, original transcript, and annotation area. Users can navigate through slides using pagination controls and save their annotations for each slide individually.

\begin{figure}[htbp]
    \centering
    \includegraphics[width=\linewidth]{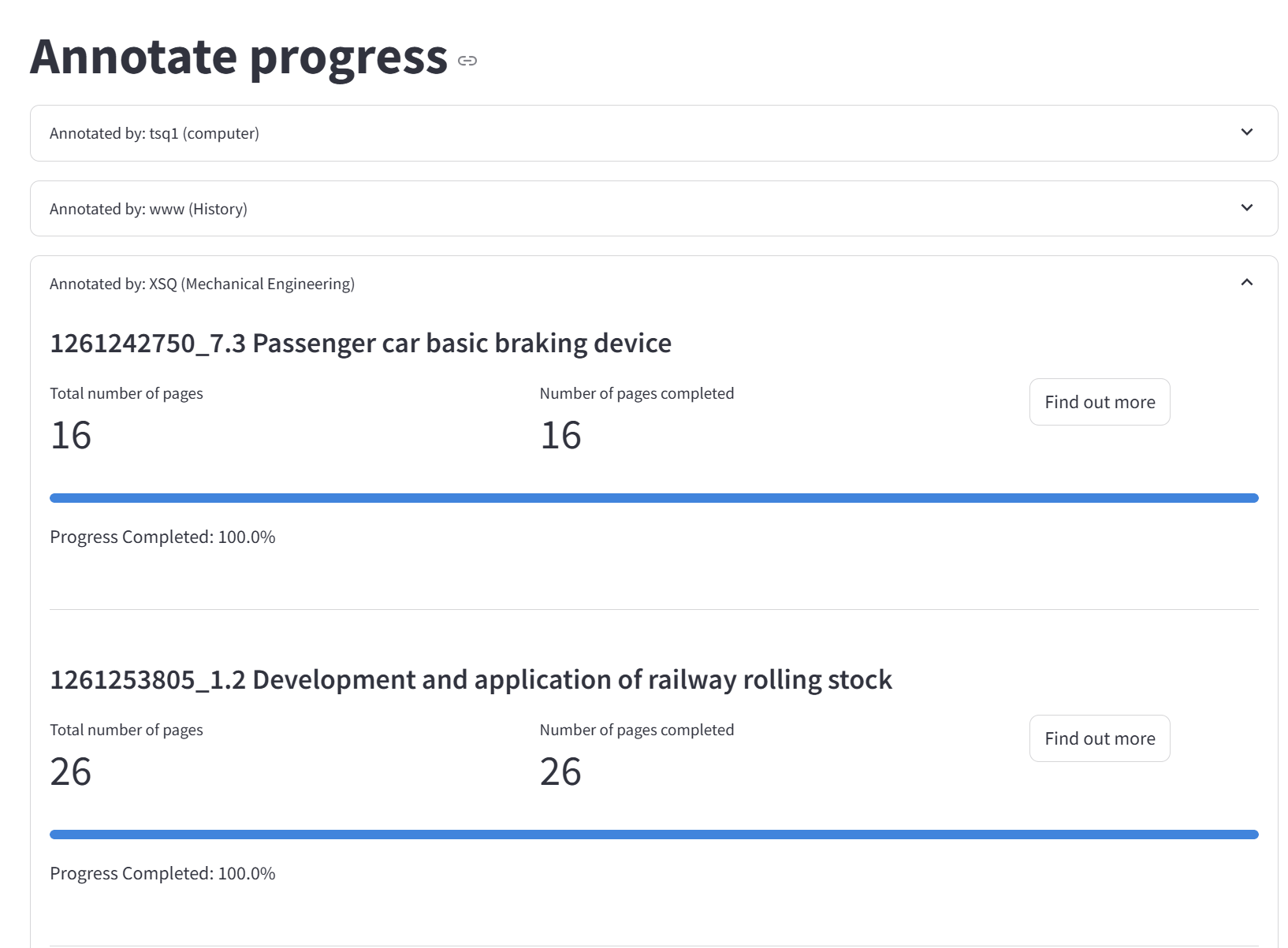}
    \caption{Screenshot of the \rednote{admin page} (progress overview). Administrators can monitor annotation progress across different majors, view detailed statistics for each course, and track overall completion status.}
    \label{fig:Annotation_Admin1}
\end{figure}

\begin{figure}[htbp]
    \centering
    \includegraphics[width=\linewidth]{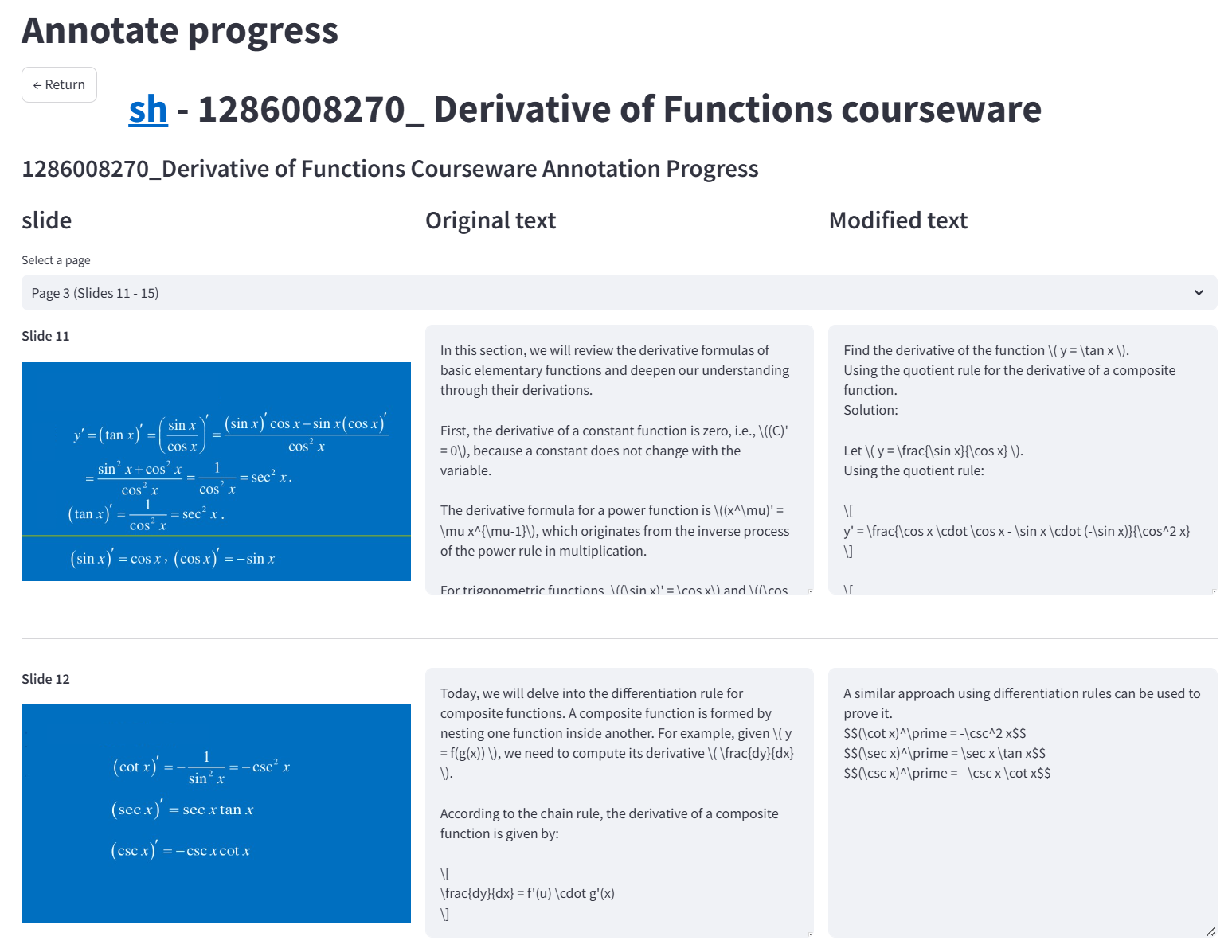}
    \caption{Screenshot of the \rednote{admin page} (review interface). Reviewers can examine both original and modified scripts side by side.}
    \label{fig:Annotation_Admin2}
\end{figure}

\noindent
\textbf{\rednote{Admin page.}} 
This page provides administrative oversight of the annotation process. As shown in Figure~\ref{fig:Annotation_Admin1}, administrators can monitor annotation progress across all majors and courses, with detailed statistics grouped by annotator's major. The interface displays progress bars and completion rates for each course, helping administrators track the overall project status. When reviewing annotations, administrators can examine both original and modified scripts side by side, as shown in Figure~\ref{fig:Annotation_Admin2}.

\clearpage
\subsection{Annotation Guidelines}
\label{sec:guide}

\noindent
\textbf{\bluenote{\underline{Guidelines for the data annotators:}}}

\begin{tcolorbox}[size=title,opacityfill=0.05,breakable]
\noindent
1. Click on ``Register'' in the left bottom part of the annotation page if this is the first time that you enter our system. You should choose your major, username and password during registration. We will assign annotation tasks according to your major. You can return to the annotation page and login via your username and password.

2. After logging in, you will see a list of PPTs that require modifications to the presentation scripts. Click on “Start Annotating” to proceed.

3. After opening a PPT, you will see the original script for each slide. These scripts are generated by AI and may have issues such as factual errors, missing information from the PPT, irrelevance to the topic, awkward phrasing, or repetitive wording. Your task is to edit the script for each slide in the "Annotation Area." After making your changes, click on "Save Annotations."

4. Note that this is a one-way system. If you press the browser's back button, the login process will restart (there are only three pages: login, list, and annotation). If you want to return to the list page from the annotation page, you can click on "Return to Course List" at the top. Remember to save your annotations promptly after completing them! Click "Save Annotations" immediately after finishing your annotations! If the PPT is not clear enough, you can use the button at the top-right corner of the image to enlarge it to full screen.

5. If you have finished marking a PPT, you can contact us for acceptance inspection. We will check each page for the following:

- Whether there are factual errors.

- Whether too much information from the PPT is missing in the script.

- Whether the content is irrelevant to the topic.

- Whether the sentences are awkward or not smooth.

- Whether there is repetitive wording.

6. Compensation:

- If every page of the PPT passes the acceptance inspection, you will receive a base payment of \texttt{80} \texttt{CNY} for each PPT.

- If a PPT fails the acceptance inspection three or more times, each subsequent failure will result in a deduction of \texttt{20} \texttt{CNY} from the payment, to compensate for the time spent by the inspector.

- If the original script of the PPT is empty on six or more pages, and the PPT passes the acceptance inspection, you will receive an additional \texttt{20} \texttt{CNY} in compensation for the time spent by the annotator in writing the script from scratch.

\textbf{After reading the above requirements, start data annotation now!}

\end{tcolorbox}

\subsection{Data Collection Cost}
We spend approximately \texttt{7,000 CNY} on human correction data collection.

\section{More Evaluation Details}
\label{sec:setup}

\subsection{Evaluation Setting}
In the Image + VLM setting, we set the generation sampling parameters to  \texttt{max\_new\_tokens}=8192. In the Caption + LLM setting, for the first model call where we use gpt-4o for generating the caption, we set \texttt{max\_new\_tokens}=1024. For the following model call where the LLM outputs the final response for the writing instruction and caption, we set \texttt{max\_new\_tokens}=8192 except for claude-3-opus, which we set \texttt{max\_new\_tokens}=4096.

\subsection{Evaluation Prompts}

\xhdr{Prompt for gpt-4o on generating caption for the \textit{Caption + LLM} setting}

\begin{tcolorbox}[size=title,opacityfill=0.05,breakable]
\noindent
Please provide a detailed and comprehensive description of the image, paying special attention to both visual elements and textual content. Consider the following aspects:

1. Main Subject(s):

   - What are the primary objects, people, or figures in the image?
   
   - Their positioning, size, and prominence
   
   - Any diagrams, charts, or graphical elements

2. Textual Content:

   - All text visible in the image, including:
   
     * Headers, titles, or captions
     
     * Labels or annotations
     
     * Body text or paragraphs
     
     * Numbers, equations, or mathematical notation
     
   - The relationship between text and visual elements

3. Visual Details:

   - Colors, lighting, and overall composition
   
   - Textures and materials visible
   
   - Any notable patterns, designs, or visual hierarchies
   
   - Quality and clarity of text/figures

4. Information Structure:

   - How information is organized (e.g., flowcharts, tables, lists)
   
   - Connections or relationships indicated by arrows or lines
   
   - Legend or key elements if present
   
   - Reading order or flow of information

5. Technical Elements:

   - Presence of graphs, charts, or scientific figures
   
   - Any coordinate systems or axes
   
   - Units of measurement or scales
   
   - Technical symbols or notation

6. Context and Purpose:

   - The apparent purpose of the image (educational, technical, decorative, etc.)
   
   - Target audience or field of study
   
   - Any relevant domain-specific context

Please provide a clear, structured description that captures both the visual and textual elements, ensuring no significant details are omitted.
\end{tcolorbox}

\xhdr{Prompt for LLMs on generating response for the \textit{Caption + LLM} setting}

\begin{tcolorbox}[size=title,opacityfill=0.05,breakable]
\noindent
Please analyze the following image captions and writing requirement carefully, then provide a detailed response that:

        1. Directly addresses the writing requirement
        
        2. Incorporates relevant details from the image captions
        
        3. Uses clear, well-structured writing
        
        4. Maintains appropriate tone and style for the context

Writing requirement: \{\textit{User Instruction}\}

Image captions: \{\textit{CAPTIONS}\}

Please provide a comprehensive response that fully satisfies the writing requirement while effectively utilizing the information from the image captions.
\end{tcolorbox}

\xhdr{Prompt for gpt-4o on scoring the quality of responses}

\begin{tcolorbox}[size=title,opacityfill=0.05,breakable]
\noindent
You are an expert in evaluating text quality. Please evaluate the quality of an AI assistant's response to a user's writing request with several corresponding images. Be as strict as possible.

You need to evaluate across the following six dimensions, with scores ranging from 1 to 5. The scoring criteria from 5 to 1 for each dimension are as follows:

1. Relevance: From content highly relevant and fully applicable to the user's request and images to completely irrelevant or inapplicable.

2. Accuracy: From content completely accurate with no factual errors or misleading information to content with numerous errors and highly misleading.

3. Coherence: From clear structure with smooth logical connections to disorganized structure with no coherence.

4. Clarity: From clear language, rich in detail, and easy to understand to confusing expression with minimal details.

5. Breadth and Depth: From both broad and deep content with a lot of information to seriously lacking breadth and depth with minimal information.

6. Reading Experience: From excellent reading experience, engaging and easy to understand content to very poor reading experience, boring and hard to understand content.

Please evaluate the quality of the following response to a user's request according to the above requirements.

<User Request>

 \{\textit{INST}\}

</User Request>

<Response>

 \{\textit{RESPONSE}\}

</Response>

Please evaluate the quality of the response. You must first provide a brief analysis of its quality, then give a comprehensive analysis with scores for each dimension. The output must strictly follow the JSON format: {"Analysis": ..., "Relevance": ..., "Accuracy": ..., "Coherence": ..., "Clarity": ..., "Breadth and Depth": ..., "Reading Experience": ...}. You do not need to consider whether the response meets the user's length requirements in your evaluation. Ensure that only one integer between 1 and 5 is output for each dimension score.
\end{tcolorbox}

\subsection{Deployment Details}
All the experiments were conducted on an Ubuntu 20.04.4 server equipped with 104 Intel Xeon(R) Platinum 8470 CPU cores, and graphic cards that contained 8 NVIDIA
A800 SXM 80GB GPUs. Besides, the CUDA version is 12.2. 
The supervised fine-tuning (SFT) phase for LongWriter-V-7B on the LongWrite-V-22k dataset took approximately six hours using 8 GPUs. For the LongWriter-V-72B model, the SFT process required 72 hours with the same GPU configuration. The Direct Preference Optimization (DPO) process for LongWriter-V-7B, using 2,844 mixed preference pairs, completed in 1.5 hours.

\clearpage
\subsection{Case Study}
\label{sec:case}

\begin{figure}[htbp]
    \centering
    \includegraphics[width=\linewidth]{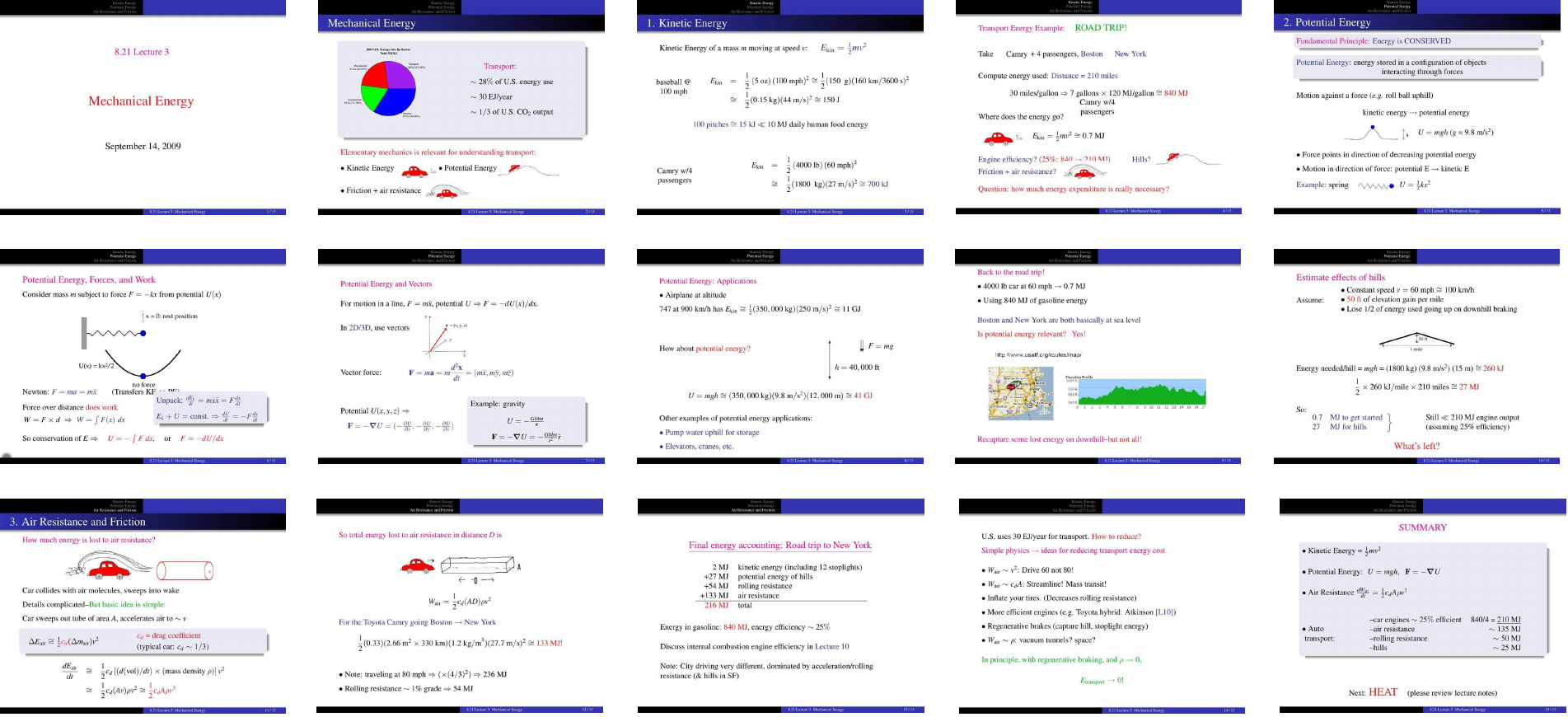}
    \caption{Input images of the example for the PPT script generation task.}
    \label{fig:input_case}
\end{figure}


To illustrate the actual outputs of our tested models, we select a case from MMLongBench-Write. As shown in Figure~\ref{fig:input_case}, this case includes 15 input images. The instruction for this case is as follows:
``Write a 3000-word lecture script based on these PPT images. The script should be suitable for a teacher to read aloud in class and must not contain any unreadable content. It should align with the teaching goals of the PPT without merely repeating its content. The script should be clear, well-structured, and possess educational and inspirational value. Please output the script directly, without additional content such as subtitles or questions, and ensure it is fully readable.''
We present the responses from GPT-4o-2024-08-06, LongWriter-V-72B, and LongWriter-V-7B-DPO below:

\noindent
\textbf{\bluenote{\underline{Response of GPT-4o-2024-08-06:}}}

\begin{tcolorbox}[size=title,opacityfill=0.05,breakable]
\noindent
Ladies and gentlemen, welcome to today's lecture on Mechanical Energy. Our focus will be on understanding the fundamental concepts of kinetic energy, potential energy, and the roles of air resistance and friction. These concepts are not just theoretical; they have practical implications in our daily lives, especially in the context of transportation.

Let's begin by considering the significance of transportation in energy consumption. In the United States, transportation accounts for approximately 28\% of total energy use, which translates to about 30 exajoules per year. This sector also contributes to one-third of the country's CO2 emissions. Understanding the mechanics behind transportation can help us devise strategies to reduce energy use and emissions.

First, let's delve into kinetic energy, which is the energy of motion. The formula for kinetic energy is \( E_{\text{kin}} = \frac{1}{2}mv^2 \), where \( m \) is mass and \( v \) is velocity. Consider a baseball thrown at 100 mph; its kinetic energy is around 150 joules. While this is a small amount of energy, it illustrates the concept effectively. Now, imagine a car like a Toyota Camry with four passengers moving at 60 mph. The kinetic energy here is approximately 700 kilojoules, demonstrating how energy scales with mass and velocity.

Let's explore a practical example: a road trip from Boston to New York, a distance of 210 miles. Suppose the car in question achieves 30 miles per gallon and consumes 7 gallons of fuel for the trip, equating to 840 megajoules of energy. However, not all this energy is used for motion. The kinetic energy required is only about 0.7 megajoules. So, where does the rest of the energy go? Factors like engine efficiency, air resistance, and friction play significant roles.

Next, we'll explore potential energy, which is stored energy based on an object's position or configuration. A fundamental principle here is energy conservation: energy cannot be created or destroyed, only transformed. Consider a ball rolling uphill; kinetic energy is converted into potential energy. The potential energy, \( U \), is given by \( U = mgh \), where \( g \) is the acceleration due to gravity, and \( h \) is height. Springs are another example, where potential energy is stored as \( U = \frac{1}{2}kx^2 \).

Now, let's consider potential energy in the context of forces and work. For a mass subject to a force, the work done is the force multiplied by the distance moved. This work results in energy transfer, such as converting potential energy to kinetic energy. In vector terms, for motion in a line, force is related to potential energy by \( F = -\frac{dU}{dx} \).

Let's apply these concepts to real-world scenarios. Consider an airplane cruising at altitude; its kinetic energy is significant. But potential energy also plays a crucial role, especially at high altitudes. Other applications include using potential energy to pump water uphill or using elevators and cranes.

Returning to our road trip example, we must consider the potential energy effects of hills. A constant speed of 60 mph with elevation gains requires additional energy. Suppose the car gains 50 feet per mile; the energy needed per hill becomes substantial. Despite this, the total energy for hills remains less than the energy output from the engine, highlighting the inefficiency of energy conversion in transportation.

Air resistance and friction are also critical factors in energy consumption. As a car moves, it collides with air molecules, creating a wake. This interaction results in energy loss, calculated using the drag coefficient. For example, a Toyota Camry traveling from Boston to New York loses a considerable amount of energy to air resistance. The drag force depends on factors like speed and surface area, and reducing these can improve efficiency.

Finally, let's summarize our findings. The total energy expenditure for our road trip includes kinetic energy, potential energy of hills, rolling resistance, and air resistance. The gasoline energy far exceeds these values, illustrating the inefficiency of internal combustion engines, which are only about 25\% efficient. This inefficiency is even more pronounced in city driving, where acceleration and rolling resistance dominate.

To reduce transport energy costs, we can apply simple physics principles. Driving at lower speeds, streamlining vehicles, inflating tires, using more efficient engines, and employing regenerative braking are effective strategies. In an ideal world, with these measures and advanced technologies like vacuum tunnels, we could drastically reduce transport energy expenditure.

In conclusion, understanding mechanical energy, particularly in transportation, provides insights into improving energy efficiency and reducing environmental impact. As we continue to innovate and apply these principles, we move closer to a more sustainable future.

Thank you for your attention. Please review your lecture notes in preparation for our next topic: Heat.

\end{tcolorbox}

\noindent
\textbf{\bluenote{\underline{Response of LongWriter-V-72B:}}}

\begin{tcolorbox}[size=title,opacityfill=0.05,breakable]
\noindent

1. Today, we delve into mechanical energy, a crucial topic in physics with broad applications, especially in transportation. The U.S. dedicates nearly 28\% of its energy to transport, amounting to about 30 exajoules annually and contributing to a third of the nation's CO2 emissions. Grasping this impact is essential for developing sustainable solutions. We'll examine how kinetic and potential energy, along with friction and air resistance, shape transport dynamics. This understanding not only illuminates our current energy use but also fuels innovation for future efficiency. As we proceed, reflect on how these principles connect to real-world challenges and inspire advancements in energy conservation.

2. Building on our introduction to mechanical energy, let's examine its pivotal role in transportation. The U.S. dedicates around 28\% of its energy to this sector, which consumes approximately 30 exajoules annually and accounts for a third of the country's CO2 emissions. This underscores the necessity for sustainable practices. To tackle these challenges, we must understand kinetic and potential energy, as well as friction and air resistance. Kinetic energy is the energy of motion, while potential energy is stored energy with the potential to perform work. Friction and air resistance are forces that impede motion, impacting energy efficiency. Grasping these concepts is crucial for optimizing transportation systems. As we delve deeper, consider how these principles can inform innovations in energy conservation and efficiency, driving progress toward a more sustainable future.

3. Let's delve into kinetic energy, the energy a moving object possesses. The formula for kinetic energy is \( E_{\text{kin}} = \frac{1}{2}mv^2 \), where \( m \) represents mass and \( v \) represents velocity. This equation reveals that kinetic energy is proportional to an object's mass and the square of its speed.

Consider a baseball traveling at 100 mph. Its kinetic energy is approximately 150 joules. Now, think about a Camry carrying four passengers moving at 60 mph; its kinetic energy jumps to around 700 kilojoules. This comparison illustrates how both mass and velocity significantly affect kinetic energy.

To further clarify, let's examine the energy needed for 100 pitches of a baseball. This amounts to about 15 kilojoules, a small fraction of the 10 megajoules of energy a person typically consumes daily through food.

These examples highlight the importance of understanding kinetic energy in various contexts, from sports to transportation. As we proceed, consider how these principles apply to real-world scenarios and how they can inform innovations in energy efficiency and sustainability.

4. Let's delve into a real-world application of energy principles by considering a road trip from Boston to New York in a Camry with four passengers. This 210-mile journey requires about 7 gallons of gasoline, equating to roughly 840 megajoules of energy. But where does this energy go?

Firstly, kinetic energy is a factor; a 4000-pound car at 60 mph possesses about 0.7 megajoules of kinetic energy. Yet, this is a small portion of the total energy used. We must also consider the engine's efficiency, typically around 25\%, meaning much of the energy is lost as heat.

Moreover, friction and air resistance play significant roles. These forces oppose the car's motion, necessitating additional energy to maintain speed. Climbing hills further increases energy demands, converting kinetic energy into potential energy.

This scenario prompts us to question the necessity of such energy expenditures. Are there ways to optimize our vehicles and driving habits to enhance efficiency? Reflect on how understanding these principles can lead to more sustainable practices in transportation. By grasping where energy is utilized and wasted, we can explore innovative solutions to reduce consumption and environmental impact.

5. Now, let's turn our attention to potential energy, a fundamental idea in physics. It represents stored energy within a system, ready to perform work. The principle of energy conservation is key here: energy cannot be created or destroyed, only transformed from one form to another.

Consider lifting an object; as it ascends, its kinetic energy converts into potential energy, expressed as \( U = mgh \), where \( m \) is mass, \( g \) is gravitational acceleration, and \( h \) is height. This potential energy can revert to kinetic energy if the object falls.

A practical example is a compressed spring. The energy you expend to compress it becomes potential energy, \( U = \frac{1}{2}kx^2 \), with \( k \) as the spring constant and \( x \) as the displacement. When released, this energy transforms back into kinetic energy.

Understanding potential energy is vital for analyzing systems and predicting motion. It enables us to calculate energy transformations and optimize processes, from elevators to water pumps. Reflect on how these concepts apply to everyday technologies and how they might inspire innovations in energy storage and efficiency.

6. Building on our understanding of potential energy, let's explore its relationship with forces and work. When an object is displaced in a potential field, like gravity or a spring, a force acts upon it. This force is derived from the potential energy gradient, expressed as \( F = -\frac{dU}{dx} \), where \( U \) is potential energy and \( x \) is position.

Work is done when a force moves an object over a distance, quantified as \( W = F \times d \). In the context of potential energy, this work equates to the change in potential energy, \( \Delta U \). For instance, lifting an object increases its gravitational potential energy, while compressing a spring increases its elastic potential energy.

These concepts are crucial in physics, as they enable us to analyze and predict system behavior. By understanding the forces and work involved, we can optimize energy use in various applications, from designing efficient machines to developing renewable energy systems. Consider how these principles apply to real-world scenarios and inspire innovations in energy efficiency and sustainability.

7. Expanding on our discussion, let's delve into potential energy in two and three dimensions using vectors. In 2D and 3D spaces, potential energy becomes a function of multiple variables, such as \( U(x, y, z) \).

The gradient of potential energy, denoted as \( \nabla U \), is a vector that points in the direction of the greatest increase in potential energy. The force acting on an object can be derived from this gradient, expressed as \( \mathbf{F} = -\nabla U \). The negative sign indicates that the force acts in the direction of decreasing potential energy.

For example, in a gravitational field, the potential energy is given by \( U = -\frac{GMm}{r} \), where \( G \) is the gravitational constant, \( M \) and \( m \) are masses, and \( r \) is the distance between them. The corresponding force is \( \mathbf{F} = -\nabla U = -\frac{GMm}{r^2} \hat{\mathbf{r}} \), which is the familiar inverse-square law of gravity.

Understanding these vector relationships is essential for analyzing complex systems, such as planetary motion or electromagnetic interactions. It allows us to predict and explain the behavior of objects under various forces, inspiring innovations in fields like robotics and aerospace engineering. Reflect on how these principles apply to real-world technologies and consider their potential to drive future advancements.

8. Let's delve into the practical applications of potential energy, beginning with aircraft. At cruising altitude, a 747 carries substantial potential energy, calculated as \( U = mgh \), where \( m \) is the aircraft's mass, \( g \) is gravitational acceleration, and \( h \) is its height. For a 747 at 40,000 feet, this equates to about 41 gigajoules.

Potential energy is also pivotal in energy storage systems. Pumped-storage hydroelectricity is a prime example, where water is pumped uphill to a reservoir during off-peak hours. When demand rises, the water is released to generate electricity, effectively converting stored potential energy into kinetic energy, which then powers turbines.

In urban settings, potential energy is harnessed in elevators and cranes. Elevators utilize counterweights to minimize energy consumption during vertical movement, while cranes employ potential energy to lift heavy loads efficiently.

These examples underscore the significance of potential energy in modern engineering and its role in sustainable energy solutions. Reflect on how these principles can inspire innovations in energy efficiency and storage, contributing to a greener future.

9. Let's revisit our road trip scenario to delve into the role of potential energy. Even though Boston and New York are at sea level, the journey between them involves elevation changes that significantly impact energy consumption.

As you ascend hills, your vehicle converts kinetic energy into potential energy, stored due to increased height. This process demands extra energy, which is then partly recovered as you descend, converting potential energy back into kinetic energy. However, not all energy is recaptured due to losses from friction and air resistance.

Understanding these dynamics is vital for optimizing vehicle efficiency. By analyzing elevation profiles, we can better predict energy requirements and improve fuel economy. This knowledge not only enhances road trip efficiency but also informs the development of smarter navigation systems and more fuel-efficient vehicles.

Consider how these principles might be applied to other modes of transport, such as trains or ships, to reduce energy usage and environmental impact. Reflect on potential innovations that could further minimize energy loss, contributing to sustainable transportation solutions.

10. To estimate the energy impact of hills during a road trip, consider a car traveling at a steady 60 mph, encountering a 50-foot elevation gain per mile. This incline necessitates additional energy to elevate the vehicle, calculated using the potential energy formula \( U = mgh \), where \( m \) is the car's mass, \( g \) is gravitational acceleration, and \( h \) is the height increase.

For a 1800 kg car, the energy required to climb each hill is approximately 260 kJ. Over a 210-mile trip, this accumulates to about 27 MJ. However, descending the hills doesn't fully recover this energy due to energy losses from braking and friction.

These calculations highlight the significant energy demands of hilly terrains. By understanding these dynamics, we can develop strategies to optimize fuel efficiency, such as using regenerative braking systems to capture energy during descents.

Reflect on how this knowledge can inform route planning and vehicle design to enhance efficiency and sustainability. Consider potential innovations that could further minimize energy loss, contributing to a greener future in transportation.

11. Now, let's delve into the impact of air resistance on vehicle energy efficiency. As a car moves, it displaces air molecules, creating drag that reduces efficiency. The energy lost to air resistance can be calculated using the formula \( \Delta E_{\text{air}} = \frac{1}{2} c_d (\Delta m_{\text{air}}) v^2 \), where \( c_d \) is the drag coefficient, \( \Delta m_{\text{air}} \) is the mass of displaced air, and \( v \) is the velocity.

For a typical car traveling at 60 mph, this results in a substantial energy loss. The loss increases with the square of the velocity, meaning higher speeds lead to significantly greater energy expenditure. For instance, a Toyota Camry traveling from Boston to New York at 60 mph loses about 133 MJ to air resistance, but this increases to 236 MJ at 80 mph.

Understanding these dynamics is crucial for enhancing vehicle efficiency. Designing more aerodynamic vehicles with lower drag coefficients can mitigate these losses. Additionally, maintaining optimal speeds can drastically improve fuel efficiency.

Reflect on how these principles can be applied to everyday driving to reduce energy consumption and environmental impact. Consider potential innovations in vehicle design and technology that could further minimize air resistance and contribute to sustainable transportation solutions.

12. Let's delve deeper into the energy lost to air resistance during a road trip. For a Toyota Camry traveling from Boston to New York at 60 mph, approximately 133 MJ is lost to air resistance. This significant amount highlights the impact of aerodynamics on fuel efficiency.

Increasing speed exacerbates this issue. At 80 mph, the energy loss rises to about 236 MJ, demonstrating that higher speeds demand more energy to overcome air resistance. This is because the energy loss is proportional to the square of the velocity.

In addition to air resistance, rolling resistance also plays a role. This is the energy lost due to tire deformation and road friction. For a typical car, rolling resistance can account for around 54 MJ over the same distance.

Understanding these factors is crucial for improving vehicle efficiency. Designing more aerodynamic vehicles and maintaining optimal speeds can significantly reduce energy loss. Keeping tires properly inflated is another simple yet effective way to minimize rolling resistance.

Reflect on how these principles can be applied to everyday driving to enhance efficiency and reduce environmental impact. Consider potential innovations in vehicle design and technology that could further mitigate these energy losses, contributing to sustainable transportation solutions.

13. Now, let's wrap up our energy analysis for the road trip from Boston to New York. We start with kinetic energy, about 2 MJ, accounting for stops along the way. Then, there's the 27 MJ for potential energy, spent on climbing hills. Rolling resistance adds another 54 MJ, and air resistance claims a significant 133 MJ. Altogether, these factors total 216 MJ of energy used.

However, the gasoline used provides 840 MJ, indicating an overall efficiency of roughly 25\%. This means only a quarter of the fuel's energy is utilized for motion, with the rest lost as heat and other inefficiencies.

This analysis underscores the inefficiencies in automobile energy usage. By grasping these components, we can pinpoint areas for improvement. For instance, enhancing aerodynamics, reducing rolling resistance, and optimizing vehicle weight can boost efficiency.

Reflect on how these insights can inform future innovations in vehicle design and technology. Consider how better understanding these dynamics can lead to more sustainable and efficient transportation solutions, reducing our environmental impact.

14. To enhance transport efficiency, we can apply fundamental physics principles. Reducing air resistance is key, as it significantly impacts energy consumption. By traveling at lower speeds—opting for 60 mph instead of 80 mph—we diminish drag and boost fuel efficiency. Streamlined vehicle designs also help cut through air more effortlessly.

Public transportation offers another efficient alternative, as buses and trains inherently possess lower per capita drag. Maintaining vehicles with properly inflated tires is crucial, as it reduces rolling resistance and conserves energy.

Innovative technologies, like regenerative braking, capture energy during braking, which can be reused, enhancing overall efficiency. This technology is particularly effective in stop-and-go traffic, maximizing energy savings.

Looking ahead, concepts such as vacuum tunnels and magnetic levitation could revolutionize transportation by drastically reducing air and rolling resistance. While these ideas are futuristic, they demonstrate the potential for significant advancements in energy efficiency.

Reflect on how these principles and innovations might be integrated into everyday travel. Consider their role in reducing environmental impact and fostering sustainable transportation solutions.

15. In conclusion, we've delved into the core concepts of mechanical energy, emphasizing kinetic and potential energy, and the effects of air resistance and friction. These elements are pivotal in understanding energy consumption in transportation.

We learned that kinetic energy, calculated as \( \frac{1}{2}mv^2 \), is essential for vehicle motion. Meanwhile, potential energy, expressed as \( mgh \), becomes significant when navigating hills or elevated terrains.

Air resistance, quantified by \( \frac{1}{2}c_d A \rho v^2 \), plays a crucial role in energy efficiency, especially at higher speeds. Minimizing this resistance can markedly enhance fuel economy.

Our energy analysis of a road trip revealed that out of the 840 MJ provided by gasoline, only 216 MJ is used for motion, highlighting the need for more efficient engines and innovative technologies.

Looking ahead, we'll explore the concept of heat and its impact on energy systems. I encourage you to review your lecture notes and reflect on how these principles can be applied to real-world challenges. Thank you for your attention, and I look forward to continuing our exploration of physics in the next session.
\end{tcolorbox}

\noindent
\textbf{\bluenote{\underline{Response of LongWriter-V-7B-DPO:}}}

\begin{tcolorbox}[size=title,opacityfill=0.05,breakable]
\noindent
**1. Kinetic Energy**

Let's continue our exploration of mechanical energy by examining kinetic energy, an essential element of transportation. The formula for kinetic energy is \( E_{\text{kin}} = \frac{1}{2}mv^2 \), where \( m \) is the mass and \( v \) is the speed. This formula reveals kinetic energy's direct relationship with both mass and velocity squared.

Imagine a baseball traveling at 100 mph. Its kinetic energy can be calculated using the formula, revealing the significant energy involved in fast-moving objects. Now, consider a car with four passengers traveling from Boston to New York. The kinetic energy involved in this journey highlights the extensive energy requirements in transportation.

Beyond calculations, understanding kinetic energy's role is vital. It affects how vehicles accelerate, brake, and navigate, influencing both safety and efficiency. Recognizing these dynamics encourages us to critically consider transportation systems' energy use and environmental impact.

As we move forward, think about how kinetic energy considerations might lead to advancements in more efficient and sustainable transport methods. Reflect on the broader implications for energy use and conservation in our daily lives.

**2. Mechanical Energy**

Transportation is a major player in the U.S. energy scene, using up about 28\% of the nation's energy resources. In 2001, transportation sector energy consumption was roughly 30 Exajoules annually, accounting for approximately one-third of the U.S. CO2 emissions. Grasping the mechanics of transportation is essential to tackle energy use and environmental concerns effectively.

Within transportation, the kinetic and potential energy involved in vehicles' movement are crucial. Kinetic energy relates to moving vehicles, impacting acceleration, speed, and braking. In contrast, potential energy arises from height or position, like when a car climbs a hill. Both types of energy are interwoven with the forces of friction and air resistance, which dissipate energy and affect efficiency.

Analyzing these energy forms helps us critically assess the efficiency and sustainability of our transportation methods. As we delve deeper, consider how a solid foundation in these mechanical principles can inspire innovations for a more energy-efficient future. Reflect on the potential to minimize environmental impact while maximizing energy use in transportation systems.

**3. Kinetic Energy**

Let's delve deeper into the concept of kinetic energy, which is crucial for understanding transportation. The formula for kinetic energy is \( E_{\text{kin}} = \frac{1}{2}mv^2 \), where \( m \) is the mass and \( v \) is the speed. This formula reveals that kinetic energy is directly proportional to the mass and the square of the velocity. 

Imagine a baseball traveling at 100 mph. Its kinetic energy can be calculated to appreciate the energy of fast-moving objects. Now, consider a car with four passengers on a journey from Boston to New York. The kinetic energy involved highlights the significant energy requirements in transportation.

Kinetic energy plays a vital role in transportation by influencing everything from fuel efficiency to emissions. Understanding these energy dynamics encourages critical thinking about how we can enhance transportation systems to make them more efficient and environmentally friendly. Reflect on how these principles might inspire innovations in the design and operation of vehicles.

For instance, think about how reducing a car's mass or modifying its speed can substantively affect its kinetic energy. How might these changes lead to greater fuel efficiency and lower emissions? Grasping these concepts not only enhances appreciation for mechanical principles but also encourages us to envision a sustainable future.

**4. Transport Energy Example: Road Trip**

Let's examine a practical example of transport energy using a road trip from Boston to New York. This journey covers about 210 miles and requires a Camry with four passengers. 

To begin, we calculate the energy consumption. The car's fuel efficiency is 30 miles per gallon, and gasoline contains roughly 120 megajoules per gallon. Thus, for this trip, the Camry will consume 7 gallons of gasoline, totaling approximately 840 megajoules of energy.

But where does all this energy go? Part of it is used to overcome kinetic energy—the energy of motion—while the car travels at speeds of around 60 miles per hour. The remaining energy is used to counteract forces like friction and air resistance, which work to slow the car down.

This example prompts us to ponder: Is the full 840 megajoules truly necessary? Are there ways to optimize energy use? By reflecting on these questions, we can explore strategies to enhance efficiency and sustainability in transportation.

Understanding energy expenditure in road trips allows us to critically evaluate transportation systems. It invites us to consider how reducing friction, optimizing speed, and improving fuel efficiency could lead to substantial energy savings. Such insights are vital for developing more sustainable transportation methods.

As we proceed, think about how these principles can be applied to other forms of transport. How might they help us craft a future with lower energy costs and a smaller environmental footprint?

**5. Potential Energy**

Now, let's turn our attention to potential energy, a crucial component of mechanical energy. Potential energy is the energy stored in an object due to its position or configuration. A classic example is a ball placed atop a hill—its height confers potential energy, which can be converted into kinetic energy as it rolls down the hill.

Grasping the conservation of energy is essential. This principle tells us that energy cannot be created or destroyed, only transformed from one form to another. As the ball rolls, its potential energy diminishes, and kinetic energy increases. At the same time, the force of gravity acts upon the ball, driving its descent and converting energy from one form to another.

Another instance is a stretched spring. The potential energy stored in the spring is due to its deformed state. When released, this energy is transferred into kinetic energy as the spring unravels.

These examples illustrate how potential energy manifests and evolves. By exploring these principles, we gain insights into energy efficiency and sustainability. Understanding these dynamics is vital for developing systems that minimize energy waste and maximize output.

As we proceed, think about applying these concepts to technological innovations. How can the principles of potential energy and conservation be harnessed to promote a more sustainable future? Reflect on these possibilities as we delve deeper into the interplay of energy forms.

**6. Potential Energy, Forces, and Work**

Let's delve into the connection between potential energy, forces, and work. Consider a mass \( m \) under the influence of a force \( F = -kx \), which is a spring force arising from a potential energy function \( U(x) \). The negative sign indicates that the force tends to reduce potential energy.

Newton's second law, \( F = ma = m\ddot{x} \), describes the force as the mass times its acceleration. This equation reveals how forces influence the motion of objects, altering their kinetic energy.

When a force acts over a distance, it performs work. The work \( W \) done by a force is given by \( W = \int F(x) \, dx \). In the case of a spring, the work performed equals the change in potential energy, \( W = -\Delta U \).

Conservation of energy is pivotal. It states that the total energy, the sum of kinetic energy \( E_k \) and potential energy \( U \), remains constant in the absence of external dissipative forces. Mathematically, \( E_k + U = \text{constant} \). This principle is fundamental in many areas of physics and engineering, as it allows us to analyze complex systems by tracking energy flow and conservation.

Reflect on how these principles can be applied to real-world scenarios, such as oscillating systems or mechanical devices. How can an understanding of forces and potential energy aid in designing more efficient systems? As we proceed, consider these applications and the broader implications for energy efficiency and sustainability.

**7. Potential Energy and Vectors**

Now, let's delve into the role of vectors in potential energy and force expressions. When objects move along a line, we use a one-dimensional vector, \( F = m\ddot{x} \), to depict the force. This vector form emphasizes the direction and magnitude of the force acting on an object.

In two or three dimensions, we employ three-dimensional vectors to fully capture the force's effect. A three-dimensional vector force is expressed as \( \mathbf{F} = m\mathbf{a} = m \frac{d^2\mathbf{x}}{dt^2} = (m\ddot{x}, m\ddot{y}, m\ddot{z}) \).

The potential energy landscape varies in complexity with dimensions. For a system with a potential energy function \( U(x, y, z) \), the force can be expressed as a gradient vector: \( \mathbf{F} = -\nabla U = \left(-\frac{\partial U}{\partial x}, -\frac{\partial U}{\partial y}, -\frac{\partial U}{\partial z}\right) \).

A key example is gravitational force, given by \( \mathbf{F} = -\nabla U = -\frac{GMm}{r^2} \hat{r} \). This equation illustrates how gravity acts to decrease potential energy, always pointing towards the center of mass.

Understanding vector representations of potential energy and forces allows us to analyze complex systems, such as planetary orbits and mechanical structures. How can the principles of vector calculus and potential energy aid in solving real-world engineering challenges? Reflect on these applications as we continue to explore the connections between these mathematical concepts and their significance in physics and beyond.

**8. Potential Energy: Applications**

Let's examine some practical applications of potential energy. Consider an airplane at high altitude. The potential energy an airplane possesses is given by \( U = mgh \), where \( m \) is the mass, \( g \) is the acceleration due to gravity, and \( h \) is the height. For a 747 traveling at 900 km/h, this potential energy can reach an impressive 11 gigajoules at an altitude of 40,000 feet. 

Potential energy is equally important on a smaller scale. Take, for example, water pumps that lift water uphill to storage tanks. This process converts electrical energy into potential energy, allowing the water to be stored until needed. Similarly, elevators and cranes utilize potential energy to lift loads against gravity, showcasing the versatility of potential energy in both engineering and everyday life.

These examples highlight the significance of potential energy in both everyday applications and large-scale industrial processes. Understanding potential energy not only helps in appreciating the efficiency of various machines but also prompts us to think creatively about energy storage and utilization in new technologies.

As we proceed, consider how these principles can be applied to sustainable energy solutions. How can the understanding of potential energy contribute to developing more efficient energy systems? Reflect on these possibilities as we continue to explore the dynamic interplay between energy forms and their applications in the world around us.

**9. Back to the Road Trip**

Let's return to our road trip scenario, where we calculated an energy expenditure of 840 megajoules for a 4000 lb car traveling at 60 mph from Boston to New York. Both cities are at sea level, so does potential energy play a role here?

Surprisingly, even at sea level, potential energy is significant due to the road's elevation profile. As you travel, the road's elevation changes can alter the car's potential energy. Climbing a hill increases potential energy, while descending allows some energy recapture through kinetic energy.

On our typical road trip, minor elevation changes occur, particularly on highways like I-95, which have noticeable elevation gains. These variations in elevation mean that potential energy is not just a theoretical concept but a practical consideration in energy accounting.

Taking potential energy into account can influence how we approach fuel efficiency. Understanding these energy dynamics encourages us to think critically about optimizing vehicle efficiency and considering the natural landscape's impact on energy use.

As we continue, think about how these principles can be applied to other real-world situations. How can recognizing potential energy help us in everyday activities or in the design of efficient transportation systems? Reflect on these possibilities and their potential to enhance our understanding of energy in the world around us.

**10. Estimate Effects of Hills**

When planning a road trip from Boston to New York, examining the impact of hills on energy use is essential. Even with a constant speed of 60 mph, elevation changes significantly affect a vehicle's energy consumption. 

Each mile gained in elevation requires approximately 50 feet of elevation gain. Assuming this scenario, 27 megajoules of energy are needed just to tackle the hills—this calculation is based on the energy cost of climbing a hill, mgh, where \( m \) is the mass, \( g \) is the acceleration due to gravity, and \( h \) is the height gained. 

This energy is partly recovered when descending, but not entirely, due to friction and other losses. Thus, some energy is recaptured on the way down, not all of it is regained. 

The implications for energy use are noteworthy. Even minor hills can result in a substantial portion of the total energy expenditure during a journey. This highlights the need to consider elevation changes when evaluating energy efficiency in driving.

These calculations encourage critical thinking about route planning and energy efficiency. As we delve deeper into energy use in transportation, think about how overcoming hills affects overall fuel consumption and what strategies could reduce this impact. How might engineering solutions or driving habits be improved to address these challenges?

**11. Air Resistance and Friction**

During a road trip, dealing with air resistance is an inevitable challenge that hampers a car's efficiency. As a vehicle moves, it interacts with air molecules, creating resistance that reduces fuel efficiency. This air resistance, or drag, is calculated using the formula \( \Delta E_{\text{air}} = \frac{1}{2} c_d (\Delta m_{\text{air}}) v^2 \), where \( c_d \) is the drag coefficient, \( \Delta m_{\text{air}} \) is the mass of air displaced, and \( v \) is the velocity of the vehicle. 

The drag coefficient varies with a vehicle's design; for typical cars, \( c_d \) is approximately 1/3. This formula reveals that energy lost to air resistance is directly linked to the vehicle's speed and the density of the air. At higher speeds, the energy loss escalates significantly.

Streamlining a car's design can substantially decrease this energy loss. By reducing the drag coefficient and minimizing surface area, vehicles can cut down on air resistance and improve fuel efficiency. This principle is evident in the aerodynamic designs of modern vehicles.

Understanding these dynamics not only highlights the importance of engineering in enhancing vehicle efficiency but also encourages us to think critically about optimizing energy use in transportation. As we proceed, consider how these principles can be applied to create more sustainable transportation methods. How might these ideas influence the future of car design and other forms of travel? Reflect on these possibilities and their potential to transform our approach to energy efficiency in the coming years.

**12. Total Energy Lost to Air Resistance**

Now, let's delve into the energy lost due to air resistance during a road trip from Boston to New York. Using the formula derived earlier, \( W_{\text{air}} = \frac{1}{2} c_d (AD) \rho v^2 \), we can estimate this energy loss. 

The drag coefficient (\( c_d \)) for a typical car is about 0.33. For a journey of 330 kilometers at a speed of 27.7 meters per second, the total energy lost to air resistance is approximately 133 megajoules. 

But how does this stack up with other energy expenditures during the trip? Consider that rolling resistance and hills each consume around 54 and 27 megajoules, respectively. Compared to the 210 megajoules total energy available from the gasoline burned, a significant portion is attributed to air resistance.

This analysis highlights the importance of reducing air resistance to enhance fuel efficiency. Streamlining a car's design, increasing tire pressure, and maintaining a moderate speed can mitigate this energy loss.

Exploring these strategies offers promising avenues to optimize vehicle efficiency. As we wrap up this lecture, think about how these insights might be applied to real-world transportation challenges. How can we harness these principles to foster more sustainable travel options? Reflect on these possibilities and their potential to transform our approach to energy efficiency in the years ahead.

**13. Final Energy Accounting**

Let's recap the various energy expenditures during our hypothetical road trip from Boston to New York. We began by accounting for kinetic energy, estimating about 2 megajoules necessary for maintaining speed and managing stoplights. 

Adding to this, the energy required to conquer hills amounts to around 27 megajoules, considering an average elevation gain of 50 feet per mile. Rolling resistance also claims approximately 54 megajoules, emulating a 1\% grade, while air resistance incurs a substantial 133 megajoules, assuming a constant speed of 60 mph. 

The total energy usage for this journey is roughly 216 megajoules. It's crucial to recognize that these calculations are based on a 25\% efficient internal combustion engine, with 840 megajoules of gasoline energy consumed during a 330 km trip.

This energy accounting highlights the impact of various factors on fuel efficiency. It directs our attention to optimization possibilities, such as enhancing engine efficiency, reducing air resistance, and exploring alternative propulsion systems.

As we analyze city driving conditions, which are heavily influenced by acceleration, rolling resistance, and stoplights, we must consider how these elements differ from long highway trips. City driving tends to consume more energy due to frequent stopping and starting.

This examination of energy use in transportation prompts us to contemplate future innovations. How can we refine our approach to energy efficiency in vehicles? Reflect on these potential advancements and their possible transformation of our relationship with energy.

**14. Reducing Transport Energy Costs**

Now, let's explore how to cut down on transport energy costs. A key piece of advice is to drive at lower speeds, as air resistance (\( W_{\text{air}} \)) is proportional to the square of the speed. Thus, choosing a speed of 60 mph instead of 80 mph can significantly cut down on energy use.

Designing vehicles to be more streamlined is another effective strategy. By reducing the drag coefficient (\( c_d \)) and the vehicle's cross-sectional area (\( A \)), we can greatly lessen air resistance. Additionally, mass transit systems offer a sustainable solution by carrying more passengers, thereby reducing the energy required per person.

Proper tire inflation is often overlooked but is crucial for reducing rolling resistance. Ensuring tires are adequately inflated can enhance fuel efficiency by decreasing the force needed to roll the tires.

Modern engines, like those in Toyota hybrids, also play a vital role. They optimize the energy used during acceleration and deceleration, improving overall efficiency. Regenerative braking systems are another feature that capture energy during braking, storing it for future use.

Looking to the future, technologies such as vacuum tunnels and space-based railways offer theoretically near-zero energy solutions. Although not yet practical, these concepts hold promise for future transportation.

Implementing these strategies can lead to substantial energy cost reductions and a more sustainable future. As you consider these ideas, think about their application in your daily life. How can you contribute to energy efficiency in transportation? Reflect on these possibilities and their potential to transform our approach to energy use in the years ahead.

**15. Lecture Summary and Next Steps**

In today's lecture, we delved into mechanical energy, starting with the fundamentals of kinetic and potential energy and their roles in transportation. We examined how factors like air resistance and friction impact energy efficiency during a road trip. Understanding these concepts is essential as they directly influence fuel consumption and emissions.

We covered strategies to reduce energy costs, such as maintaining optimal speed, enhancing vehicle design, and utilizing advanced engine technologies. Our aim is to spark practical applications in everyday driving and future innovations.

Next, we will explore the topic of heat, a critical area in understanding energy efficiency and thermodynamics. This new journey will build on the mechanical energy principles we just discussed, promising to be both engaging and thought-provoking. Review your notes and reflect on the connections between these subjects as you prepare for the upcoming lecture. Your participation and engagement will be key as we continue to explore these vital topics.

\end{tcolorbox}

\end{document}